\documentclass[lettersize,journal]{IEEEtran}
\usepackage{graphicx}
\usepackage{amsmath}
\usepackage{amssymb}
\usepackage{framed}
\usepackage{array}
\usepackage{subfig}
\usepackage{ragged2e}
\usepackage{setspace}
\usepackage{makecell}
\usepackage{color}

\usepackage{booktabs} 
\usepackage{multirow}
\usepackage[normalem]{ulem}
\usepackage{tabularray}
\usepackage[pagebackref=true,colorlinks,bookmarks=false,citecolor=blue,linkcolor=blue,urlcolor=blue]{hyperref}

\usepackage{xurl}
\usepackage{nicematrix}
\usepackage[table]{xcolor}
\definecolor{lightcyan}{RGB}{240,248,255}

\ifCLASSOPTIONcompsoc
  \usepackage[nocompress]{cite}
\else
  \usepackage{cite}
\fi

\ifCLASSINFOpdf
\else
\fi

\begin{document}
\title{Towards High Fidelity Face Swapping: A Comprehensive Survey and New Benchmark}

\author{ % <-this % stops a space
        Qi~Li,~\IEEEmembership{}
        Weining~Wang,~\IEEEmembership{}
        Shuangjun Du,~\IEEEmembership{}
        Bo Peng,~\IEEEmembership{}
        Jing Dong,~\IEEEmembership{}
        Kun Wang,~\IEEEmembership{}
        Zhenan Sun,~\IEEEmembership{} \\
        and Ming-Hsuan Yang~\IEEEmembership{}

\IEEEcompsocitemizethanks{
\IEEEcompsocthanksitem 
Qi Li, Shuangjun Du, Bo Peng, Jing Dong and Zhenan Sun are with the New Laboratory of Pattern Recognition (NLPR), State Key Laboratory of Multimodal Artificial Intelligence Systems (MAIS), Institute of Automation, Chinese Academy of Sciences, Beijing, 100190, China, and are also with the School of Artificial Intelligence, University of Chinese Academy of Sciences, Beijing 100049, China (email: qli@nlpr.ia.ac.cn;  dushuangjun2024@ia.ac.cn; bo.peng@nlpr.ia.ac.cn; jdong@nlpr.ia.ac.cn; znsun@nlpr.ia.ac.cn).

\IEEEcompsocthanksitem 
Weining Wang is with the Laboratory of Cognition and Decision Intelligence for Complex Systems, Institute of Automation, Chinese Academy of Sciences, Beijing, 100190, China (email: weining.wang@nlpr.ia.ac.cn).

\IEEEcompsocthanksitem 
Kun Wang is with Nanyang Technological University, Singapore 639798, Singapore (email: wang.kun@ntu.edu.sg).

\IEEEcompsocthanksitem 
Ming-Hsuan Yang is with the Department of Computer Science and
Engineering, University of California, Merced, CA, 95340, USA (email:mhyang@ucmerced.edu), and Department of Computer Science and Engineering, Yonsei University, Korea. 

}
}

\markboth{IEEE TRANSACTIONS ON PATTERN ANALYSIS AND MACHINE INTELLIGENCE}%
{Shell \MakeLowercase{\textit{et al.}}: A Sample Article Using IEEEtran.cls for IEEE Journals}

%{Shell \MakeLowercase{\textit{et al.}}: Bare Demo of IEEEtran.cls for Computer Society Journals}

\IEEEtitleabstractindextext{%
\begin{abstract}
\justifying
Face swapping has witnessed significant progress in recent years, largely driven by advances in deep generative models such as GANs and diffusion models.
Despite these advances, existing methods remain fragmented across different paradigms, and their evaluation is highly inconsistent due to the lack of standardized datasets and protocols. Moreover, prior surveys primarily focus on broader deepfake generation or detection, leaving face swapping insufficiently studied as a standalone problem. In this paper, we present a comprehensive survey and benchmark for face swapping. We provide a structured review of existing methods, organizing them into five major paradigms and systematically analyzing their design principles, strengths, and limitations. To enable fair and controlled evaluation, we introduce CASIA FaceSwapping, a high-quality benchmark with balanced demographic distributions and explicit attribute variations, and establish standardized protocols to assess the robustness of different face swapping methods. Extensive experiments on representative approaches yield new insights into the performance characteristics and limitations of current techniques. Overall, our work provides a unified perspective and a principled evaluation framework to facilitate the development of more robust and controllable face swapping methods. 
More results can be found at \href{https://github.com/CASIA-NLPRAI/face-swapping-survey}{https://github.com/CASIA-NLPRAI/face-swapping-survey}.
\end{abstract}

\begin{IEEEkeywords}
face swapping survey, face swapping benchmark, face swapping evaluation 
\end{IEEEkeywords}}

\maketitle
\IEEEdisplaynontitleabstractindextext
\IEEEpeerreviewmaketitle

\section{Introduction}
\label{sec:introduction}

\IEEEPARstart{F}{ace} swapping aims to transfer the identity of a source face onto a target while preserving target-specific attributes such as pose, expression, illumination, and background. Beyond its direct applications in privacy protection, digital entertainment, and virtual avatars, face swapping also serves as a representative testbed for studying identity–attribute disentanglement in generative modeling. It is not merely an application-oriented image manipulation task, but also a concrete instance of a broader research problem: how to separate, control, and recombine semantic factors in realistic visual generation.

Early face swapping methods are primarily based on encoder--decoder architectures, which model facial geometry or learn latent representations for identity and attributes. Subsequently, GAN-based methods have advanced the field by leveraging powerful generative priors and disentangled latent spaces, enabling high-fidelity and controllable synthesis. More recently, diffusion-based methods have emerged as a new paradigm, benefiting from strong generative capacity and flexible conditioning, and achieving state-of-the-art performance in realism and identity preservation.

Despite rapid progress, a systematic understanding of face swapping methods remains limited. As shown in Table~\ref{tab:Tab1}, existing surveys predominantly focus on broader deepfake generation or detection, rather than treating face swapping as a standalone problem with distinct methodological challenges. Moreover, the relationships between architectural design choices (e.g., latent manipulation, spatial control, and conditional guidance) and their empirical behaviors remain insufficiently understood.
At the same time, the empirical evaluation of face swapping methods is far from standardized. Most existing works rely on heterogeneous datasets and experimental settings, many of which are originally designed for related tasks such as deepfake detection rather than face swapping. These datasets often suffer from identity overlap between training and testing splits, limited demographic diversity, and the lack of carefully designed evaluation protocols that isolate specific factors such as ethnicity and attribute variation. As a result, current evaluations often fail to rigorously assess model generalization, fairness, and robustness, potentially leading to misleading conclusions.

\begin{table*}[t]
\small
\centering
\caption{Summary of representative surveys on deepfake generation and detection.}
\vspace{-0.5em}
\rowcolors{2}{lightcyan}{white}
\resizebox{\textwidth}{!}{
\begin{tabular}{m{0.30\textwidth} m{0.20\textwidth} m{0.50\textwidth}}
\toprule
\textbf{Title} & \textbf{Publication} & \textbf{Descriptions} \\
\midrule
The Creation and Detection of Deepfakes: A Survey~\cite{mirsky2021creation}  
& ACM Computing Surveys 2021  
& Early overview of deepfake creation and detection techniques, without dedicated analysis of face swapping. \\
Deep Learning for Deepfakes Creation and Detection: A Survey~\cite{nguyen2022deep}  
& Computer Vision and Image Understanding 2022  
& Covers both generation and detection methods across modalities, but lacks fine-grained categorization for face swapping. \\
Countering Malicious Deepfakes: Survey, Battleground, and Horizon~\cite{juefei2022countering}  
& International Journal of Computer Vision 2022  
& A comprehensive analysis of DeepFake generation, detection, and evasion, including a taxonomy of methods and a depiction of the adversarial dynamics between attackers and defenders. \\
Deepfakes Generation and Detection: State-of-the-Art, Open Challenges, Countermeasures, and Way Forward~\cite{masood2023deepfakes}  
& Applied Intelligence 2023  
& A detailed analysis of existing tools and machine learning-based approaches for deepfake generation, as well as the methodologies used to detect such manipulations for both audio and visual deepfakes. \\
Deepfake Detection: A Comprehensive Survey from the Reliability Perspective~\cite{wang2024deepfake}  
& ACM Computing Surveys 2024  
& A comprehensive review of existing DeepFake detection studies from the perspective of reliability, interpretability, and robustness. \\
Deepfake Generation and Detection: A Benchmark and Survey~\cite{pei2024deepfake}  
& ACM Computing Surveys 2026  
& A comprehensive review of recent advancements in deepfake generation and detection. \\
\bottomrule
\end{tabular}
}
\label{tab:Tab1}
\end{table*}

To address these limitations, we present a comprehensive survey and benchmark for face swapping. 
We propose a unified taxonomy that categorizes existing methods by core design principles, including 3D model-based, autoencoder-based, GAN-based, StyleGAN-based, and diffusion-based approaches, with fine-grained distinctions reflecting their dominant mechanisms. 
We further introduce CASIA FaceSwapping, a high-quality benchmark designed for fair and fine-grained evaluation, featuring 4K videos, balanced demographics, and controlled attribute variations. Furthermore, we establish standardized protocols to evaluate three key aspects: baseline performance, cross-ethnicity generalization, and robustness to attribute variations. 
Extensive experiments on representative methods provide insights into their performance characteristics and limitations. The main contributions of this paper are:
\begin{itemize}
	\item We present the first comprehensive survey on face swapping, introducing a unified taxonomy that organizes methods by design principles.

	\item We introduce CASIA FaceSwapping, the first large-scale benchmark with 4K video resolution, balanced demographic composition, and controlled attribute variations, which is specifically designed for face swapping evaluation. 

   \item We establish three standardized evaluation protocols, namely Normal, Cross-ethnicity, and Cross-attribute, to facilitate fair and factorized analysis of the performance of different methods.

    \item We conduct extensive experiments and analyses on representative approaches, providing new empirical insights into their generalization ability and robustness, which can inform future research in this area.
\end{itemize}

\section{Preliminaries}
\label{sec-preliminaries}

\subsection{DeepFake Algorithm}
\begin{figure}[tp]
  \begin{center}
  \includegraphics[width=0.95\linewidth]{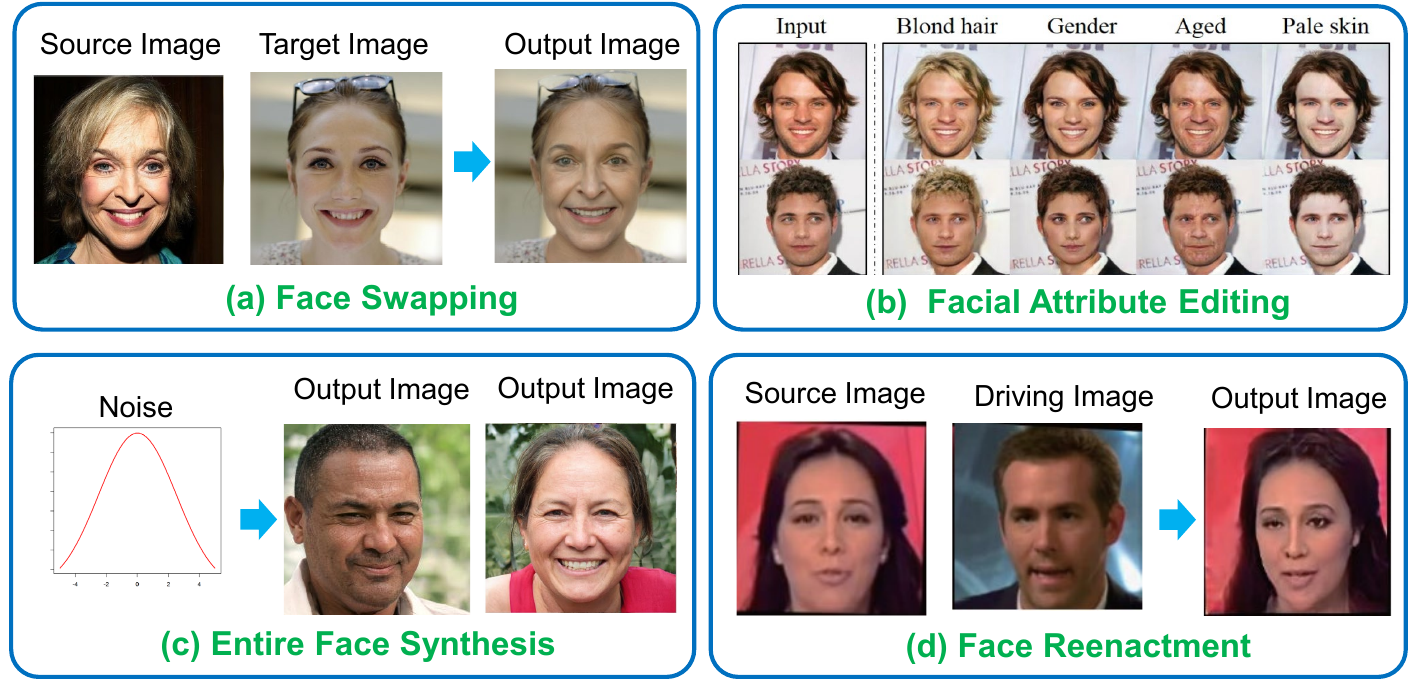}
  \end{center}
  \vspace{-0.2cm}'
  \caption{Sample results obtained by (a) face swapping method (MegaFS)~\cite{zhu2021one},  (b) facial attribute editing method (Stargan)~\cite{choi2018stargan}, (c) entire face synthesis (StyleGAN)~\cite{karras2019style}, and (d) face reenactment method (DCG-GAN)~\cite{liu2022semantic}. 
  \vspace{-0.2cm}
  }
\label{fig:intro_FAM_samples}
\end{figure}

DeepFake technology refers to the creation and manipulation of facial appearance using deep generative methods, typically categorized into four types: face swapping, facial attribute manipulation, entire face synthesis, and face reenactment. Figure~\ref{fig:intro_FAM_samples} illustrates representative examples of these categories. 
Apart from face swapping, the remaining concepts are defined as follows. Facial attribute editing refers to modifying generic facial components or soft biometrics, excluding identity and expression. Entire face synthesis generates a new face image from noise. Face reenactment animates a source face using the pose and expression of a driving image while preserving identity.

\subsection{Baseline Models}

To better structure the taxonomy, we first review several fundamental techniques that serve as building blocks for face swapping methods.

\vspace{1mm} \noindent \textbf{3D Morphable Model (3DMM).} 3DMM is one of the most successful models for representing faces in 3D space. It defines a linear subspace for shape and texture using principal component analysis (PCA), and can be formulated as:
\begin{equation}
\begin{array}{l}
{S_{3D}} = {{\bar S}_{3D}} + {{B}_{id}}{\alpha _{id}} + {{B}_{exp }}{\beta _{exp }},\\
{T_{3D}} = {{\bar T}_{3D}} + {{B}_{tex}}{\alpha _{tex}},
\end{array}
\end{equation}
where $S_{3D}$ and $T_{3D}$ denote the 3D shape and texture, ${\bar S}_{3D}$ and ${\bar T}_{3D}$ are the mean shape and texture, ${B}_{id}$, ${B}_{exp}$, and ${B}_{tex}$ are PCA bases for identity, expression, and texture, respectively, and $\alpha_{id}$, $\beta_{exp}$, and $\alpha_{tex}$ are the corresponding coefficients.

\vspace{1mm} \noindent \textbf{Encoder-Decoder Network.} An encoder-decoder network consists of an encoder $E_e$ and a decoder $D_e$, designed to map an input $x$ to a latent representation and reconstruct it. The encoder projects the input space $X$ to a latent space $Z_e$: $z_e = E_e(x) \in Z_e$, while the decoder maps $Z_e$ to the output space $\hat{X}$: $\hat{x} = D_e(z_e) \in \hat{X}$. When the output is trained to reconstruct the input, the model is referred to as an autoencoder. The reconstruction objective is:
$$
\mathop{\min}\limits_{E_e, D_e} \; \mathbb{E}_{x \in X} \left[ \| x - \hat{x} \|_2 \right],
$$
where $\|\cdot\|_2$ denotes the $\ell_2$ norm. Variants include contrastive and variational autoencoders.

\vspace{1mm} \noindent \textbf{Generative Adversarial Network (GAN).} A typical GAN consists of a generator $G_g$ and a discriminator $D_g$. The generator maps a latent code $z_g \in Z_g$ to an image $\hat{x} = G_g(z_g)$, while the discriminator distinguishes generated samples from real data $x \in X$. The two networks are trained adversarially via:
\begin{align}
\begin{split}
\mathop{\min}\limits_{G_g} \mathop{\max}\limits_{D_g} \; & \mathbb{E}_{x \in X} \left[ \log D_g(x) \right] \\
& + \mathbb{E}_{z_g \in Z_g} \left[ \log \left(1 - D_g(G_g(z_g)) \right) \right].
\end{split}
\end{align}
To improve image quality and resolution, numerous variants have been proposed. Among them, StyleGAN and its extensions~\cite{karras2019style,karras2020analyzing,karras2021alias} introduce architectural innovations distinct from standard GANs and are discussed separately.

\vspace{1mm} \noindent \textbf{StyleGAN.} StyleGAN~\cite{karras2019style} introduces a style-based generator. It starts from a learned constant input and modulates ``style'' at each convolution layer using a latent code ${z_s} \in {Z_s}$. The latent code is mapped to an intermediate space $W$ via a mapping network ${f_s}$: ${w} = {f_s}({z_s}) \in W$. The code $w$ is then passed through affine transformations to produce style codes, which, together with injected noise, modulate feature maps at different layers to control attributes at multiple scales. 
To improve reconstruction, the $W^+$ space extends $W$ by using distinct latent codes for each layer, enhancing fidelity~\cite{abdal2019image2stylegan,abdal2020image2stylegan++}. The $S$ space further defines channel-wise style coefficients, improving spatial disentanglement for image editing~\cite{wu2021stylespace,liu2022towards}..

\vspace{1mm} \noindent \textbf{Diffusion Models.} Diffusion models are a class of probabilistic generative models that can progressively degrades data samples by injecting noise and then learn to reverse the process for generation new samples. We briefly introduce the denoising diffusion probabilistic model (DDPM)~\cite{ho2020denoising}.  DDPM consists of two Markov processes: a forward process to transform data samples into a simple prior distribution by adding Gaussian noise, and a reverse process to reverse the forward process by learning transition kernels parameterized by deep neural networks. 
Given an original data sample ${x_0}$ that follows distribution $p\left( {{x_0}} \right)$, the noised versions ${x_1},{x_2},...,{x_T}$ are obtained by incrementally transform the data distribution into a tractable prior distribution, i.e.,  
$p\left( {{x_t}|{x_{t - 1}}} \right) = \mathcal{N}\left( {{x_t};\sqrt {1 - {\beta _t}} {x_{t - 1}},{\beta _t}{\rm I}} \right)$, where $T$ is the number of diffusion steps, ${\beta _t} \in \left[ {0,1} \right]$ is a hyperparameter representing the variance schedule across diffusion steps, $\mathcal{N}\left( {x;u,\sigma } \right)$ denotes the normal distribution with mean $u$ and variance $\sigma$. 
For the reverse process, given the prior distribution $p\left( {{x_T}} \right) = \mathcal{N}\left( {{x_T};0,{\rm I}} \right)$, the new samples can be generated as: $p\left( {{x_{t - 1}}|{x_t}} \right) = \mathcal{N}\left( {{x_{t - 1}};\mu \left( {{x_t},t} \right),\Sigma \left( {{x_t},t} \right)} \right)$. To approximate these steps, a neural network can be trained to predict the mean and the covariance. In practice, the neural network can also be trained to predict the noise from the image and compute the mean and covariance.

\section{Review of Face Swapping Methods}
\label{sec-review-face-swapping}

In this section, we review representative face swapping algorithms under a unified taxonomy. Existing approaches can be broadly grouped into five paradigms: 3DMM-based, autoencoder-based, GAN-based, StyleGAN-based, and diffusion-based methods. These paradigms differ in their underlying representations and in how they disentangle and recombine identity and attributes. For clarity, we first introduce the notations used throughout this section. Let the source image be denoted as $x_s$ and the target image as $x_t$. Face swapping aims to synthesize an output ${x_{s \to t}}$ that preserves the identity of $x_s$ while inheriting the attributes (e.g., pose, expression, illumination, and background) of $x_t$. We then review these paradigms and discuss their key design choices, strengths, and limitations.

\subsection{3DMM-based Methods}
3DMM-based methods use the 3D morphable model~\cite{blanz2023morphable} to estimate parameters for swapping. A typical pipeline reconstructs a 3D face and scene parameters from a single 2D image, and then swaps the corresponding parameters to achieve face swapping.

\begin{figure}[t]
  \begin{center}
  \includegraphics[width=1.0\linewidth, height=0.6\linewidth]{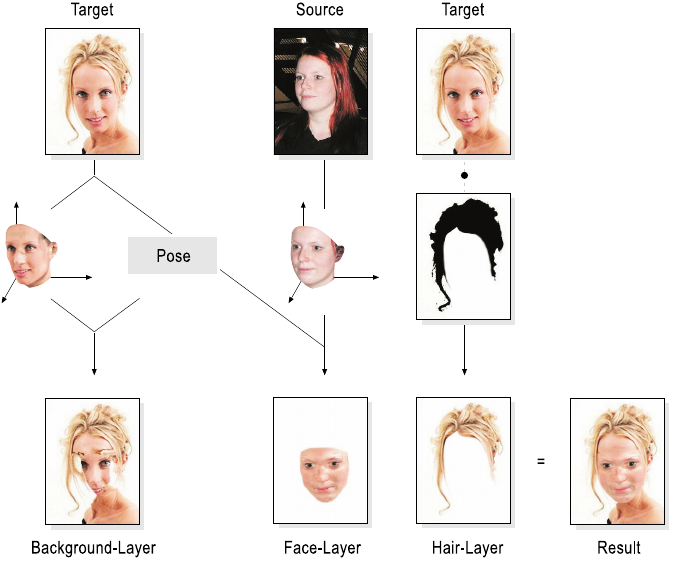}
  \end{center}
  \vspace{-0.2cm}
  \caption{3DMM-based face swapping framework of~\cite{blanz2004exchanging}.
  \vspace{-0.2cm}
  }
\label{fig:exchanging-faces}
\end{figure}

Building on this pipeline, Blanz et al.~\cite{blanz2004exchanging} propose a system that represents 3D faces in a vector space of shapes and textures. As shown in Figure~\ref{fig:exchanging-faces}, the result is synthesized using three layers: face, hair, and background. The face layer is reconstructed from the source with rendering parameters from the target, the hair layer is copied from the target, and the background is generated via background continuation~\cite{blanz2003reanimating}. A limitation is the need for manual steps, such as annotating facial landmarks and hairlines. 
Cheng et al.~\cite{cheng20093d} reduce user intervention but still require similar pose, expression, and illumination. To relax these constraints, Lin et al.~\cite{lin2014pose} propose a pose-free method that reconstructs the source in 3D, estimates the target pose, and renders a virtual source image for blending. Lin et al.~\cite{lin2012face} further address these issues using personalized 3D head models.

Modern 3DMM-based methods increasingly integrate neural networks to improve robustness and efficiency. Rather than relying solely on iterative optimization, recent approaches use regression networks to estimate 3DMM parameters, often combined with differentiable rendering for end-to-end learning. This paradigm improves stability in real-world scenarios. Many methods further incorporate neural refinement to recover high-frequency details not captured by the low-dimensional 3DMM space.

Peng et al.~\cite{peng2021unified} use BFM to decompose a face into pose, shape, and expression via optimization, and recombine these factors with a U-Net-like generator for swapping. Nirkin et al.~\cite{nirkin2018face} estimate 3D shape from landmarks and segment facial regions, generating results via blending. HifiFace~\cite{hififace} regresses 3DMM coefficients and extracts identity features using a pre-trained recognition network, which are fused with attribute features via AdaIN~\cite{huang2017arbitrary}. Otto et al.~\cite{otto2022learning} propose a shared encoder with identity-specific decoders that produce geometry and texture, rendered via a differentiable renderer. Li et al.~\cite{li20233d} map inputs into the EG3D~\cite{chan2022efficient} latent space, disentangle and swap codes, and generate high-fidelity, 3D-consistent results, albeit with higher inference cost.

\subsection{Autoencoder-based Methods}

\begin{figure}[t]
  \begin{center}
  \includegraphics[width=1.0\linewidth]{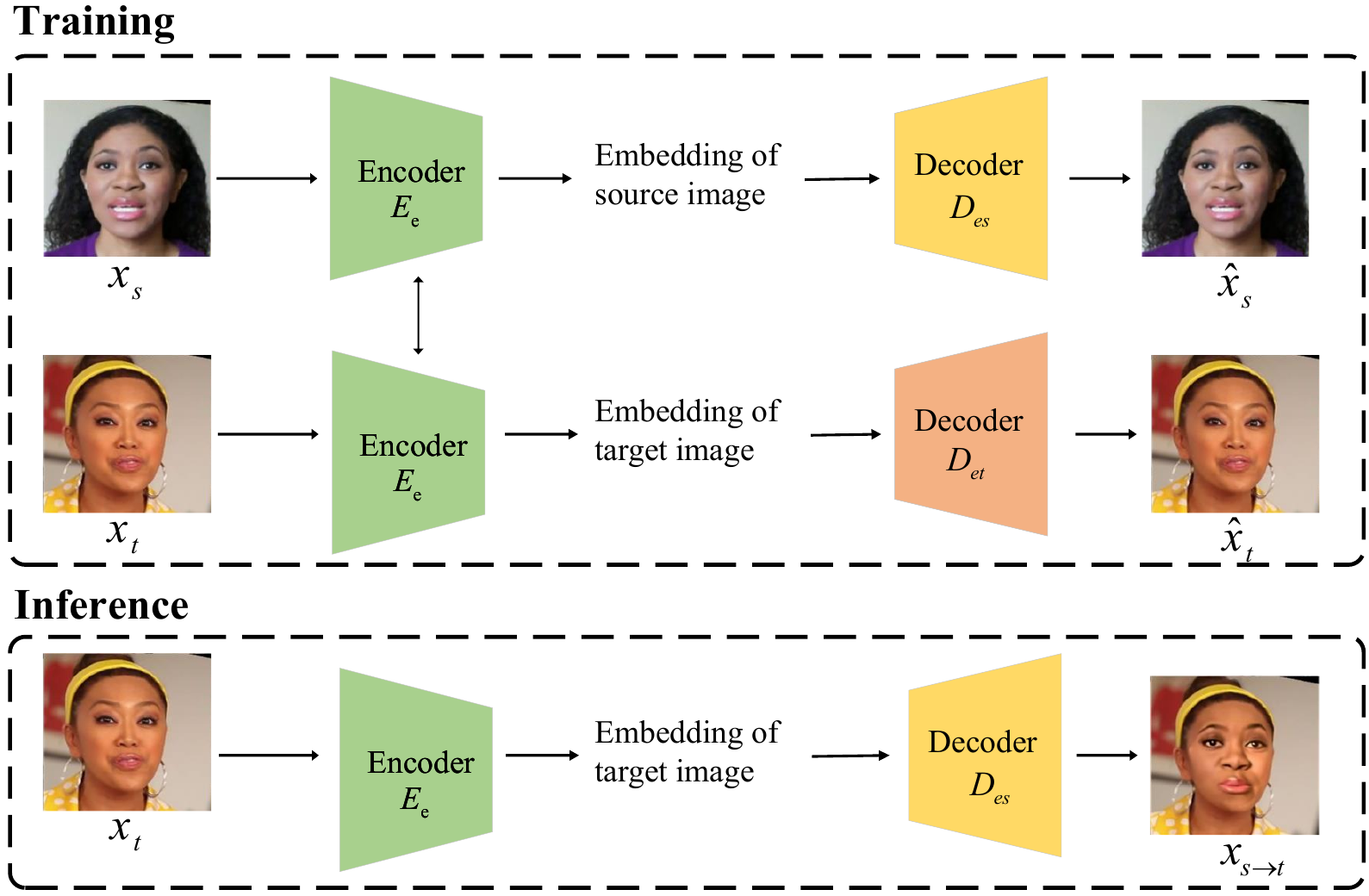}
  \end{center}
  \vspace{-0.2cm}
  \caption{An example of a typical autoencoder-based face swapping method.
  \vspace{-0.2cm}
  }
\label{fig:auto-encoder-deepfake-sample}
\end{figure}

The original DeepFake technique~\cite{DeepFake-url} uses a shared encoder $E_e$ and two decoders $D_{es}$ and $D_{et}$. 
During training, $E_e$ extracts common features, while $D_{es}$ and $D_{et}$ capture identity-specific features of the source and target. 
Thus, both $x_s$ and $x_t$ are encoded into a shared latent space, while each decoder reconstructs its corresponding identity. 
At inference, the target image $x_t$ is passed through $E_e$ and $D_{es}$ to produce the swapped result: ${D_{es}}({E_e}({x_t})) = {x_{s \to t}}$. 
Figure~\ref{fig:auto-encoder-deepfake-sample} shows a typical example. 
Note that this method is subject-specific and must be trained separately for each pair of subjects.

Inspired by style transfer, Korshunova et al.~\cite{korshunova2017fast} treat identity as style and facial dynamics as content, using a multi-scale texture network with VGG-19 content and style losses. Wilson et al.~\cite{wilson2023introducing} note that standard image losses poorly capture eye regions and introduce a gaze constraint via a pre-trained estimation network to improve realism. Liu et al.~\cite{liu2023high} propose S2Swap, which disentangles identity and attributes at both global and local levels within an encoder--decoder framework. Zhu et al.~\cite{zhu2024stableswap} present StableSwap, a reversible autoencoder that maps images into a shared latent space and enables stable training and flexible manipulation, with multi-stage identity injection using global embeddings, 3D attributes, and landmarks.

\subsection{GAN-based Methods}

GAN-based methods are a major line of research in face swapping due to their strong generative capability and flexible feature learning. FaceSwap-GAN~\cite{FaceswapGAN-url} extends the DeepFakes framework~\cite{DeepFake-url} by incorporating adversarial and perceptual losses to improve realism. Since then, many GAN-based approaches have been proposed, with the core challenge of disentangling identity and attributes and recombining them for synthesis. From a modeling perspective, these methods can be broadly categorized into image translation-based and style transfer-based approaches, depending on whether style modulation is explicitly used. Recent works also emphasize efficiency and deployability, targeting resource-constrained platforms via lightweight architectures and accelerated inference.

\subsubsection{Image Translation-based Methods}

This line of work formulates face swapping as an image-to-image translation problem, often relying on intermediate operations such as segmentation and blending. A typical pipeline segments facial regions in source and target images, generates a swapped face with source identity and target attributes, and blends it into the target background. While intuitive, such pipelines are often not fully end-to-end.

\begin{figure}[t]
  \begin{center}
  \includegraphics[width=1.0\linewidth]{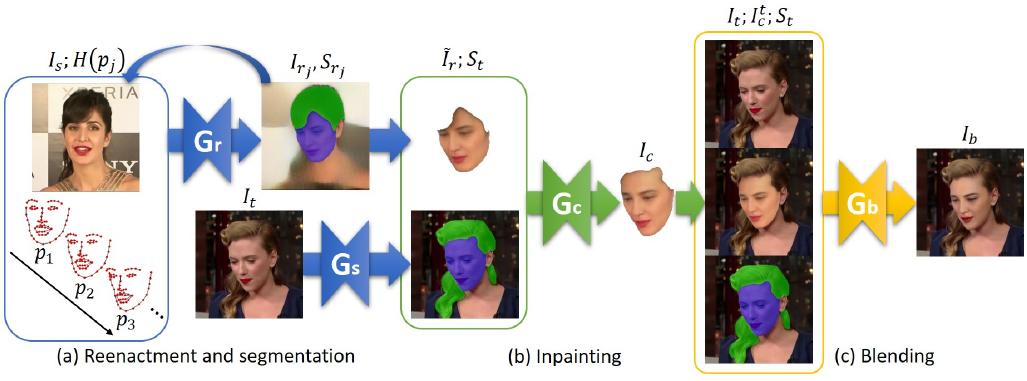}
  \end{center}
  \vspace{-0.2cm}
  \caption{The flowchart of FSGAN~\cite{nirkin2019fsgan}.
  \vspace{-0.2cm}
  }
\label{fig:fsgan}
\end{figure}

A representative image translation-based method is FSGAN~\cite{nirkin2019fsgan}, as shown in Figure~\ref{fig:fsgan}. FSGAN is an early subject-agnostic framework for face swapping and reenactment. It first generates a reenacted face using sparse 3D landmark tracking and a segmentation mask, then adopts a modular pipeline with 3D reconstruction, reenactment, and inpainting networks to align pose and blend results. By disentangling identity, expression, and pose, FSGAN enables flexible swapping. However, sparse landmarks limit fine-grained expression modeling. FSGAN v2~\cite{nirkin2022fsganv2} replaces them with a landmark transformer for more accurate reconstruction and introduces improved inpainting with facial priors, along with additional processing to reduce artifacts.

HeadSwapper~\cite{shu2022few} follows a similar pipeline by reenacting the source head and compositing it into the target via alignment and blending modules, ensuring consistent expressions and seamless integration. RSGAN~\cite{natsume2018rsgan} separates face and hair using variational autoencoders and recombines them via a GAN-based composer, achieving robust results but limited by resolution.

With the rise of vision transformers~\cite{dosovitskiy2021an}, recent methods incorporate attention mechanisms. TransFS~\cite{cao2023transfs} uses a Swin Transformer-based encoder and identity generators to reconstruct high-resolution faces, combined with warping, color correction, and blending modules for improved controllability. Face Transformer~\cite{cui2023face} learns semantic relationships between source and target regions for identity transfer, but may produce color inconsistencies between face and neck. 

\subsubsection{Style Transfer-based Methods}

\begin{figure}[t]
  \begin{center}
  \includegraphics[width=1.0\linewidth, height=0.7\linewidth]{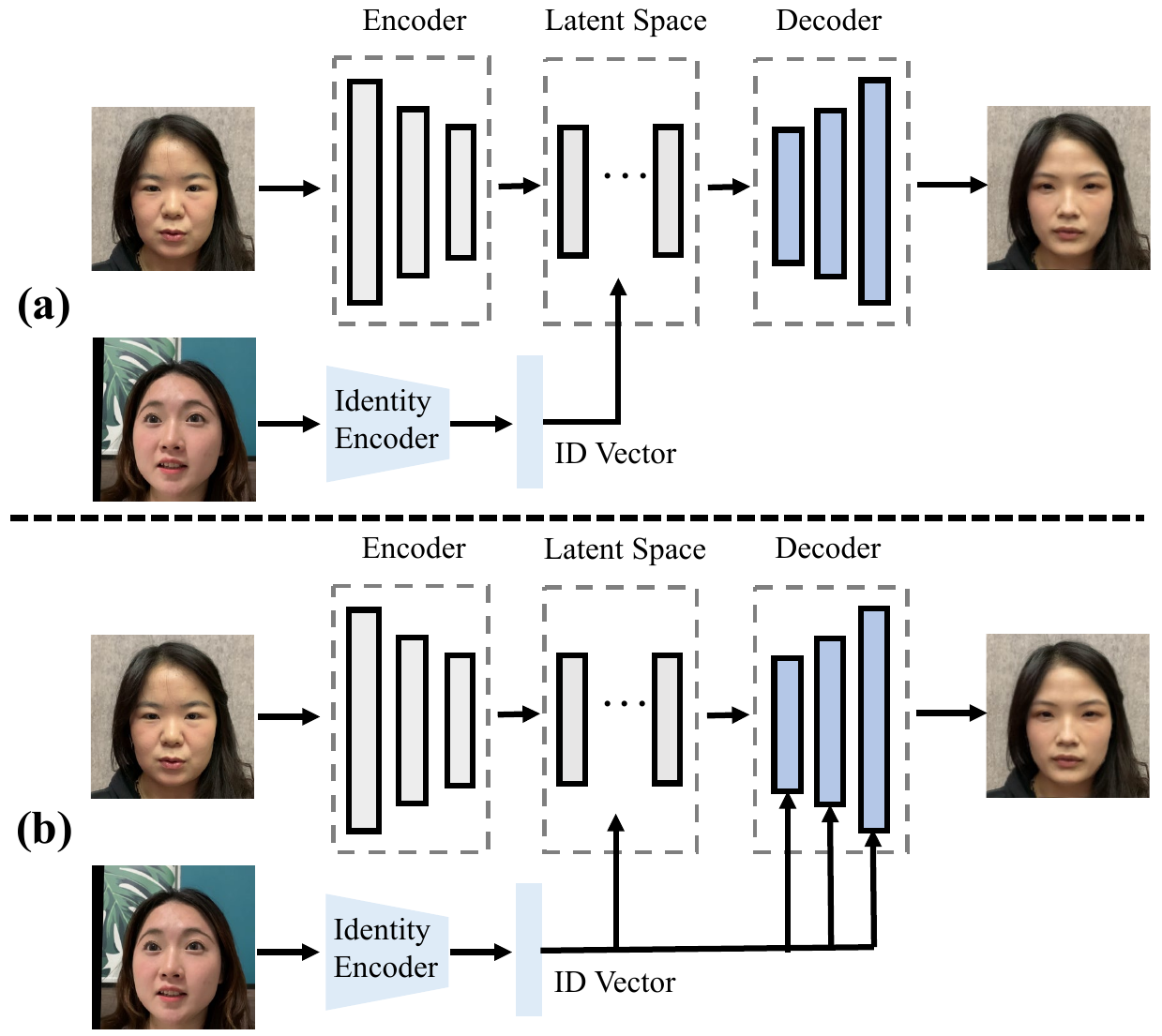}
  \end{center}
  \vspace{-0.2cm}
  \caption{Illustration of style transfer–based methods. (a) Latent-space style transfer performs identity injection primarily in the latent space (bottleneck). (b) Multi-scale style transfer applies identity modulation at both latent space and decoder stages.
  \vspace{-0.2cm}
  }
\label{fig:style-transfer}
\end{figure}

In contrast, style transfer-based methods integrate identity and attributes via explicit style modulation. These methods typically include an encoder, a decoder, and an identity injection mechanism. Depending on whether modulation is applied only at the bottleneck or across multiple decoding stages, they can be categorized into latent-space and multi-scale approaches. As shown in Figure~\ref{fig:style-transfer}, latent-space methods inject identity at the bottleneck (e.g., via AdaIN~\cite{huang2017arbitrary} or attention~\cite{vaswani2017attention}), while multi-scale methods perform identity injection throughout the decoder, enabling finer control and more consistent identity transfer.

\vspace{1mm} \noindent \textbf{Latent Space Style Transfer-based Methods.}
SimSwap~\cite{chen2020simswap} extracts identity embeddings from the source using a pretrained recognition network and injects them into a U-Net generator via AdaIN-based residual blocks, enabling feature-wise modulation to transfer identity while preserving target attributes. This alignment of identity-related statistics yields high-fidelity results. SimSwap++~\cite{chen2024simswap++} improves efficiency by introducing Conditional Dynamic Convolution (CD-Conv), which integrates identity more effectively than AdaIN, especially in compact models. UniFace~\cite{xu2022designing} adopts an attention-based approach, generating feature displacement fields to warp source features and injecting identity via attention modules.

These methods improve inner-face transfer but often suffer from shape misalignment. FlowFace~\cite{zeng2023flowface} addresses this by adding a face reshaping network that estimates semantic flow for geometric alignment, followed by cross-attention fusion to integrate identity and attributes. FlowFace++~\cite{zhang2023flowface++} further enhances this design by pretraining the reshaping network and using it as a discriminator. SelfSwapper~\cite{lee2024selfswapper} proposes a self-supervised framework that disentangles identity from non-identity attributes via self-reconstruction, introducing perforation confusion and mesh scaling to reduce bias, along with a skin color encoder and neutral-albedo rendering for improved identity preservation.

\vspace{1mm} \noindent \textbf{Multi-scale Style Transfer-based Methods.} FaceShifter~\cite{li2020advancing} is a representative method that extracts source identity via a pretrained recognition network and encodes target attributes using a multi-scale U-Net. An adaptive attentional denormalization (AAD) generator injects identity and attributes at multiple scales, with a refinement network handling occlusions. However, FaceShifter imposes weak attribute constraints and may introduce attributes. RSFace~\cite{yang2023rsface} improves this by introducing an attribute matching loss and synthesizing a reenacted face to better preserve identity. Paste2Game~\cite{zeng2022paste} enforces expression consistency and uses dual recognition networks to reduce bias.

To address limitations of fixed identity encoders, FaceSwapper~\cite{li2024learning} introduces learnable identity and attribute encoders for more flexible injection. BlendFace~\cite{shiohara2023blendface} trains identity encoders on blended images to focus on identity-relevant features, while InfoSwap~\cite{gao2021information} employs an information bottleneck to disentangle identity and non-identity features.

Beyond encoder design, several works improve disentanglement through complementary strategies. CSCS~\cite{huang2024identity} uses proxy supervision with paired data for pixel-level identity consistency. FaceDancer~\cite{rosberg2023facedancer} employs mapping networks and attention-based fusion without explicit segmentation. WSC-Swap~\cite{ren2023reinforced} removes skip connections and introduces identity removal losses with dual encoders for better disentanglement. Smooth-Swap~\cite{kim2022smooth} instead constructs a smooth identity embedding space via contrastive learning, achieving competitive performance with a simpler U-Net architecture and basic losses.

\subsubsection{Efficient Face Swapping}
The high computational complexity of many existing face swapping methods limits their deployment on resource-constrained devices, motivating the development of more efficient architectures. FastSwap~\cite{yoo2023fastswap} proposes a lightweight single-stage framework that separately extracts identity and pose features and fuses them in the decoder via adaptive normalization, enabling efficient identity transfer. MobileFaceSwap~\cite{xu2022mobilefaceswap} further targets mobile video scenarios by introducing an Identity Injection Module and an Identity-Aware Dynamic Network, where identity information is injected through weight modulation and background consistency is maintained via a weakly semantic fusion mechanism. Extending to ultra-low-power settings, PhiNet-GAN~\cite{ancilotto2023phinet} adopts a simplified encoder–decoder design for many-to-one face swapping, allowing generic identity encoding while preserving target attributes on microcontrollers.
Overall, these methods demonstrate that efficient face swapping can be achieved through a combination of lightweight architectural design, adaptive feature fusion, and training optimization techniques, making them suitable for real-time  applications on edge devices.

\subsection{StyleGAN-based Methods}

Unlike pixel-space blending or encoder-decoder reconstruction, StyleGAN-based face swapping methods operate by manipulating latent representations within a pretrained generative prior.  They have become a major paradigm for high-fidelity face swapping due to the strong semantic structure of StyleGAN latent spaces. The core challenge is to transfer source identity while preserving target-specific attributes during latent generation. From the perspective of how the latent structure is exploited, existing methods can be broadly categorized into three groups: global latent manipulation methods, region-aware modulation methods, and hierarchical factor modeling methods. 
The first group performs swapping through global editing,  or recombination of latent codes in spaces such as $W+$, offering simplicity with pretrained StyleGAN generators, but often suffering from limited controllability due to the entanglement of identity and attributes. The second group introduces spatially localized control through facial masks, or attention modules, enabling more precise identity injection into specific facial regions and improving the preservation of background. The third group further exploits the coarse-to-fine generation hierarchy of StyleGAN and incorporates more structured modeling of facial factors, allowing identity–attribute recombination at different semantic levels, and thus generally achieving stronger controllability and higher fidelity.

\subsubsection{Global Latent Manipulation Methods}

MegaFS~\cite{zhu2021one} is a representative method in this category, which is the first method to achieve one-shot face swapping at megapixel resolution. It performs GAN inversion to obtain hierarchical facial representations by predicting multiple latent codes together with the constant input of StyleGAN.   A dedicated face transfer module is then introduced to assemble these latent codes, and the final swapped image is synthesized using a pre-trained StyleGAN generator. 
Following a similar paradigm, LatentSwap~\cite{choi2024latentswap} adopts a dual-branch encoder to extract identity and attribute codes from the source and target images, respectively. A learnable mapper fuses these codes into a new latent representation, enabling flexible identity transfer without retraining the generator.  
Furthermore, SCLSS~\cite{yang2023high} enhances the inversion stage by integrating convolutional and transformer-based encoders, allowing more effective modeling of both structural and texture information. During the swapping stage, an identity feature response mechanism is introduced to select appropriate feature layers for identity–attribute decoupling, leading to improved high-fidelity synthesis.
FSALL~\cite{lin2023end} proposes an end-to-end framework for high-resolution face swapping via adaptive latent representation learning. It introduces a multi-task dual-space encoder to separately model pose and facial attributes, together with an adaptive latent code swapping module for more accurate attribute transfer. Unlike MegaFS and SCLSS, which rely on face-mask to preserve the target background, FSALL integrates facial perception and blending into a unified pipeline, thereby reducing post-processing artifacts. 
FSLSD~\cite{xu2022high} further exploits the progressive nature of StyleGAN by drawing structure and appearance cues from shallow and deep layers, respectively, enabling finer disentanglement of identity and attribute.

\subsubsection{Region-aware Modulation Methods}

StyleFace~\cite{luo2022styleface} is a representative example in this category. It maps source identity features extracted by a pre-trained face recognition network into the $W+$ space, thereby injecting identity information into the generated face. To better preserve target-specific attributes, it further employs an adaptive attribute extractor that predicts masks for non-identity regions and uses them to selectively filter attribute information at different scales.
E4S~\cite{liu2023fine} further strengthens region-level control by first generating a reenacted face and then using a mask-guided multi-scale encoder to capture the shape and texture of each facial region based on facial segmentation.By swapping the shape and texture of corresponding regions, the final face is generated using StyleGAN. This fine-grained control, facilitated by facial segmentation, allows E4S to adeptly handle facial details and occlusions. 
RAFSwap~\cite{xu2022region} also performs region-sensitive identity injection through attention, but differs from segmentation-only methods by additionally modeling global identity cues, such as wrinkles, through global average pooling and multilayer perceptrons. 
It also predicts a soft identity-relevant mask to guide the integration between the swapped face and the target background.

\subsubsection{Hierarchical Factor Modeling Methods}
StyleSwap~\cite{xu2022styleswap} is a representative hierarchical method that injects source identity into the pretrained StyleGAN at multiple stages while simultaneously predicting a facial contour mask to preserve the target background. Instead of relying solely on external segmentation for localized editing, it exploits the multi-layer generation process for structured identity transfer. Going beyond hierarchical injection, ControlFace~\cite{zhang2024controlface} further introduces factor-aware modeling by explicitly disentangling identity into structure and texture components using two 3D autoencoders together with a 2D face recognition network. These representations are then hierarchically injected into the $W+$ latent space through dedicated feature mappers aligned with coarse-to-fine semantic levels, enabling multiple controllable swapping modes.
Compared with prior methods, ControlFace achieves higher identity fidelity and enables fine-grained user control, while maintaining high-resolution output and semantic consistency.
 StyleIPSB~\cite{jiang2023styleipsb} also moves toward factor-aware modeling by combining 3D morphable model guidance with latent manipulation in the $W+$ space, where different groups of style codes are associated with pose, expression, and illumination. In contrast, MFIM~\cite{na2022mfim} performs face swapping in the $S$ space under 3D supervision and assigns low- and high-resolution style codes to global and local facial information, respectively. To preserve spatial details, MFIM not only uses style codes with channel dimensions but also extracts style maps with spatial dimensions. These style maps serve as noise maps for the pre-trained StyleGAN, providing fine details to the swapped results.

\subsection{Diffusion-based Methods}
With the development of diffusion models~\cite{ho2020denoising}, face swapping has increasingly been formulated as a conditional generation problem in which the synthesized image must simultaneously preserve source identity and target-specific attributes. Compared with earlier generative paradigms, diffusion models provide stronger generation capability and finer control, making them a promising framework for high-fidelity face swapping. Rather than differing mainly in sampling strategies, existing diffusion-based methods are  mainly distinguished by how conditioning information is introduced and how facial priors are incorporated into the generation process. They can be broadly categorized into three groups: expert-guided diffusion methods, inpainting-based diffusion methods, and structured prior-based diffusion methods.
Expert-guided diffusion methods introduce external expert models, such as face recognition and face parsing,  to provide identity-related or attribute-related guidance during generation. Inpainting-based diffusion methods formulate face swapping as a conditional inpainting problem, where the target context is preserved and the face region is selectively regenerated under source identity guidance. Structured prior-based diffusion methods incorporate explicit facial priors into the diffusion process, enabling more precise identity-attribute recombination and stronger controllability.

\subsubsection{Expert-guided Diffusion Methods}
DiffFace~\cite{kim2022diffface} is a representative method in this category, which is among the first to apply diffusion models to face swapping. It introduces three facial expert models, including a face recognition model, a face parsing model, and a gaze estimation network, which guide the diffusion process from the perspectives of source identity, target attributes, and gaze consistency, respectively. In addition, a target-preserving blending strategy is employed to adaptively maintain the target background, enabling high-fidelity face swapping results. 
Face-Adapter~\cite{han2024face} further extends this line by introducing a plug-and-play adapter for pretrained diffusion models. Without modifying the backbone U-Net, it disentangles and injects identity, pose, expression, and attribute control signals through a spatial condition generator, an identity encoder, and an attribute controller. By unifying face swapping and reenactment as a conditional generation problem, Face-Adapter achieves fine-grained control with relatively low training cost. Although these methods offer strong flexibility in integrating multiple expert signals, their performance may be constrained by the robustness and compatibility of the external expert models.

\subsubsection{Inpainting-based Diffusion Methods}
DiffSwap~\cite{zhao2023diffswap} is a representative inpainting-based method. It feeds source identity features, target landmarks, and source facial region features into the diffusion model to generate swapped results, while treating face swapping as a controllable inpainting process. To better align facial geometry, it further extracts 3D parameters and replaces the target shape parameter with that of the source to construct landmark guidance. REFace~\cite{baliah2025realistic} also reformulates face swapping as a conditional inpainting task, but shifts more complexity to training in order to improve inference efficiency. It enhances identity transfer and visual fidelity through multi-step DDIM sampling, disentangled CLIP features for pose and expression preservation, and a mask shuffling strategy. Compared with earlier diffusion-based approaches that perform identity conditioning and blending mainly at inference time, REFace provides a more efficient and stable inpainting-based solution. Overall, these methods offer a clear separation between preserved and regenerated regions, making them particularly suitable for scenarios that require strong background preservation and localized editing control.

\subsubsection{Structured Prior-based Diffusion Methods}

UniFace++~\cite{xu2025uniface++} is a representative method in this category. Unlike its predecessor UniFace, which is built on a GAN-based architecture, UniFace++ adopts a diffusion-based generation process that significantly improves training stability and visual fidelity. It revisits the unified framework for face reenactment and face swapping by leveraging disentangled and interpretable facial representations together with a diffusion-based generative backbone. 
The method encodes faces into a compact latent space structured around explicit 3D-aware semantic priors, including identity, expression, pose, and illumination, and recombines them through dedicated rendering modules for controllable generation. HS-diffusion~\cite{wang2022hs} addresses the related task of head swapping by conditioning latent diffusion on semantic layouts, enabling coherent generation beyond the facial region. 

Recent video-based diffusion methods further strengthen structured prior modeling for temporally consistent face swapping. VividFace~\cite{shao2024vividface} proposes an image-video hybrid diffusion framework with a unified latent space learned by a pseudo-3D VAE. To maintain realism, it integrates three conditioning signals: face masks for localized inpainting, 3DMM reconstructions to guide pose and expression without leaking appearance, and disentangled face-encoder features for identity, texture, and attributes. HiFiVFS~\cite{chen2024hifivfs} extends diffusion-based face swapping into a true video-to-video framework by building upon stable video diffusion. It introduces complementary modules for fine-grained attribute learning and detailed identity learning, significantly improving long-range temporal stability. DynamicFace~\cite{wang2025dynamicface} introduces a diffusion-based framework that leverages composable 3D facial priors to achieve high-quality and temporally consistent video face swapping. The method decompose the face into four explicit conditions: background, shape-aware normal map, expression-related landmark, and identity-removed UV texture map. These priors are spatially aligned via 3D reconstruction and injected into the diffusion process through dedicated condition encoders, thereby enabling high-quality and temporally consistent video face swapping. CanonSwap~\cite{luo2025canonswap} further proposes a canonical-space modulation framework that decouples motion and appearance by performing identity modulation in a canonical pose space before projecting results back to the original dynamic space.

\section{Datasets and Evaluation Metrics}
\label{sec-datasets-metrics}

\subsection{Datasets}

\begin{table*}[!htbp]
\scriptsize
\renewcommand{\arraystretch}{1.05}
\setlength{\tabcolsep}{3pt}
\centering
\caption{Representative datasets used for evaluating face swapping algorithms.}
\label{tab:Tab2}
\begin{tabular}{lccccccccc}
\toprule
\textbf{Method} & 
\textbf{\#Videos} & 
\textbf{\#Frames} & 
\textbf{\#Subjects} & 
\textbf{Resolution} & 
\textbf{Pose} & 
\textbf{Expression} & 
\textbf{\makecell{Colorful\\illumination}} & 
\textbf{\makecell{Variations\\per Subject}} & 
\textbf{Race} \\
\midrule
UADFV~\cite{yang2019exposing} 
& 49 & 17.3K & 49 & 294×500 & Near-frontal & Natural & No & No 
& \makecell{Predominantly\\Caucasian} \\

\rowcolor{lightcyan}
Deepfake TIMIT~\cite{korshunov2018deepfakes} 
& 320 & 34.0K & 32 & 128×128 & Near-frontal & Natural & No & No 
& \makecell{Predominantly\\Caucasian} \\

FaceForensics++~\cite{rossler2019faceforensics++} 
& 1,000 & 509.9K & 885 & $<$1920×1080 & Near-frontal & Natural & No & No 
& \makecell{Predominantly\\Caucasian} \\

\rowcolor{lightcyan}
Celeb-DF~\cite{li2020celeb} 
& 590 & 225.4K & 59 & \makecell{256×256\\(face region)} & Near-frontal & Natural & No & No 
& \makecell{Asian: 5.1\%\\African: 6.8\%\\Caucasian: 88.1\%} \\

DeeperForensics~\cite{jiang2020deeperforensics} 
& 50,000 & 12.6M & 100 & 1920×1080 
& \makecell{Multi-view\\(yaw/pitch/roll)} 
& \makecell{Eight\\expressions} 
& No & No & Multi-racial \\

\rowcolor{lightcyan}
DFDC~\cite{dolhansky2020deepfake} 
& 20,000 & 40M & 960 & 1920×1080 & Near-frontal & Natural & No & No 
& Multi-racial \\

CASIA FaceSwapping 
& 2,582 & 2.83M & 1,291 & 2160×3840 
& \makecell{Multi-view\\(yaw/pitch/roll)} 
& \makecell{Seven\\expressions} 
& Yes 
& \makecell{One normal\\One attribute variation} 
& \makecell{Asian: 34.6\%\\African: 28.8\%\\Caucasian: 36.6\%} \\

\bottomrule
\end{tabular}
\end{table*}

In this paper, we primarily focus on video-based datasets for face swapping evaluation. While image-based datasets such as CelebA~\cite{liu2015deep} and CelebA-HQ~\cite{karras2018progressive} have been widely used for facial attribute analysis and generative modeling, they remain static image collections and thus cannot fully capture the temporal dynamics required in face swapping. Video datasets, on the other hand, provide continuous sequences that better reflect real-world scenarios. There are several video datasets used for evaluating different face swapping algorithms. Most of these datasets are  initially designed for facial forgery detection, which we introduced as follows.

\vspace{1mm} \noindent \textbf{UADFV:} The UADFV dataset~\cite{yang2019exposing} is an early video-based dataset for facial forgery detection. 
It contains real YouTube videos and corresponding fake videos generated using a DNN model with FakeApp~\cite{fakeapp-url}. It comprises 49 videos with about 17,300 frames. Each video has a typical resolution of 294 $\times$ 500 pixels.

\vspace{1mm} \noindent \textbf{Deepfake TIMIT:} The Deepfake TIMIT~\cite{korshunov2018deepfakes} is another standard dataset for deepfake detection introduced in 2018.  It consists of videos in which faces are swapped using the open-source faceswap-GAN~\cite{FaceswapGAN-url}, a model derived from the original DeepFake algorithm~\cite{original2017deepfake}.
The dataset is created by manually selecting 32 subjects from the VidTIMIT dataset~\cite{sanderson2009multi}, and training two models to generate manipulated videos at different quality levels, with resolutions up to 128 × 128 pixels. In total, 640 fake videos are crafted based on the 320 real videos.

\vspace{1mm} \noindent \textbf{FaceForensics++:} The FaceForensics++ (FF++) dataset~\cite{rossler2019faceforensics++} is a large-scale video dataset originally designed for facial forgery detection.  It contains 1,000 original Youtube video sequences that have been manipulated with five automated face modification methods: DeepFakes~\cite{DeepFake-url}, Face2Face~\cite{thies2016face2face}, FaceSwap~\cite{FaceSwap-url},
NeuralTextures~\cite{thies2019deferred} and FaceShifter~\cite{li2020advancing}. 
FF++ is  one of the most widely adopted benchmarks for evaluating different  face swapping methods.

\vspace{1mm} \noindent \textbf{Celeb-DF:}  The Celeb-DF dataset~\cite{li2020celeb} consists of 590 real videos collected from YouTube interviews of 59 subjects. 5,693 manipulated videos are generated using an improved DeepFake synthesis algorithm. Compared with earlier datasets, Celeb-DF provides higher-quality videos and effectively reduces the visible artifacts commonly present in previous benchmarks.

\vspace{1mm} \noindent \textbf{DeeperForensics:} The DeeperForensics dataset~\cite{jiang2020deeperforensics} is a large-scale dataset  for real-world face forgery detection. It consists of 60,000 videos with 17.6 million frames in total, including 50,000 original collected videos and 10,000 manipulated videos.  The source data are collected from 100 paid actors under controlled conditions with diverse poses, expressions, and illuminations.

\vspace{1mm} \noindent \textbf{DFDC:} The Facebook DeepFake  detection challenge (DFDC) dataset~\cite{dolhansky2020deepfake} is one of  the largest publicly available face forgery dataset. It includes 960 paid actors and actresses speaking in a variety of settings. It has 128,154 DeepFake videos created based on about 20,000 real videos, whose average recording time is 68.8 seconds.

A summary of existing video-based evaluation datasets are shown in Table~\ref{tab:Tab2}.   From the table we can see that existing datasets have greatly contributed to the development of face forgery detection and manipulation research. They vary in scale, resolution, data diversity, and manipulation methods, ranging from early small-scale collections (e.g., UADFV, DeepfakeTIMIT) to large-scale benchmarks with millions of frames (e.g., FaceForensics++, DFDC, DeeperForensics). 
While these datasets provide valuable resources, they exhibits several limitations when repurposed for evaluating face swapping algorithms.
First, the dataset does not provide detailed annotations regarding facial attributes (e.g., pose, expression, lighting), nor does it support controlled evaluations across these factors. 
Second, these datasets primarily focuses on detection rather than identity preservation or attribute fidelity, and the original videos are limited in diversity, particularly in terms of demographic distribution.  It does not allow for fine-grained assessment of swapping quality across demographic boundaries.
Third,  the identity distribution across those videos is not clearly controlled, many videos share the same identities, which can lead to biased or inflated identity retrieval scores.  To illustrate,  pristine video sequences in the FaceForensics++ dataset are downloaded from the internet and some videos may belong to the same identity. For instance, videos \#043 and \#343 show the same person, Vladimir Putin, and
videos \#179, \#183 and \#826 contain the same person, Barack Obama.
 These limitations motivate the construction of our new CASIA FaceSwapping dataset, which specifically targets high-quality face swapping evaluation with balanced demographics, controlled variations, and standardized testing protocols.

\subsection{Evaluation Metrics}
\label{subsec:evaluation-metrics}

The widely used evaluation metrics can be summarized from two aspects: the accuracy of face swapping and the realism of the swapped results. 
Recall that face swapping involves transferring the face from a source image to a target image while preserving the pose, expression, lighting, and other attributes of the target. Common evaluation metrics include identity retrieval accuracy (ID retrieval), pose error, and expression error, which respectively assess the preservation of identity, pose, and expression in the swapped results.
In addition, Fr$\boldsymbol{\acute{e}}$chet Inception Distance (FID)~\cite{heusel2017gans} is also commonly employed to measure the realism of the generated images. Furthermore, we also introduces two  metrics to evaluate the temporal consistency of different face swapping methods.

\vspace{1mm} \noindent \textbf{ID Retrieval}: ID retrieval measures the identity preservation ability of different face swapping algorithms. It first extracts the identity features via the pre-trained face recognition model~\cite{wang2018cosface}.  For each swapped face image, the cosine similarity between its identity feature and those of all source images is computed. The source image with the highest similarity is retrieved, and the retrieval is considered correct if it shares the same identity as the swapped image. ID retrieval is calculated as the average accuracy of all such retrievals.

\vspace{1mm} \noindent \textbf{ID Similarity}: ID similarity also  measures the identity preservation ability of different face swapping algorithms. The identity features are extracted via the pre-trained face recognition model~\cite{wang2018cosface}.  Then cosine
similarity between identity features of the swapped faces and the corresponding source faces is computed as ID similarity. 

\vspace{1mm} \noindent \textbf{Pose Error}: Pose error measures how well different face swapping algorithms preserve the original head pose. Specifically, head poses are estimated using the HopeNet model~\cite{doosti2020hope}, and the $\ell_2$ distance between the pose vectors of the swapped and target face images is computed. The final pose error is reported as the average of these distances across all test samples.
  
\vspace{1mm} \noindent \textbf{Expression Error}: Expression error quantifies the ability of face swapping algorithms to preserve facial expressions. Expression features are extracted using a 3D face model~\cite{deng2019accurate}, and the $\ell_2$ distance is calculated between the expression vectors of the swapped and target face images. The final expression error is reported as the mean of these distances across the dataset.

\vspace{1mm} \noindent \textbf{FID}: FID evaluates image quality by measuring the discrepancy between the feature distributions of generated and real images. 
Specifically, it computes the Fr$\boldsymbol{\acute{e}}$chet distance between multivariate Gaussian distributions fitted to the Inception-v3 feature representations~\cite{szegedy2016rethinking} of the two image sets. A lower FID score indicates a closer alignment between the distributions, reflecting higher visual fidelity. Compared to earlier metrics such as the Structural Similarity Index Measure (SSIM) and the Inception Score (IS), FID has demonstrated greater stability and stronger correlation with human judgment.

\vspace{1mm} \noindent \textbf{Temporal Consistency Metrics:} 
Inspired by VBench~\cite{huang2023vbench}, we adopt two metrics, subject consistency and background consistency, to assess temporal stability of swapped results. Subject consistency measures identity preservation across frames using DINO features~\cite{dino2021}, while background consistency evaluates scene stability using CLIP features~\cite{clip2021}. Both are computed as the average cosine similarity between each frame and the first and preceding frames.

\section{Benchmarks}
\label{sec-benchmarks}

We first describe the data acquisition procedure of  CASIA FaceSwapping dataset. Then, we examine the defining characteristics of our benchmarks and delineate the principal differences between our dataset and related public datasets.
 
\subsection{Data Statics}

\begin{figure}
  \begin{center}
  \includegraphics[width=1.0\linewidth]{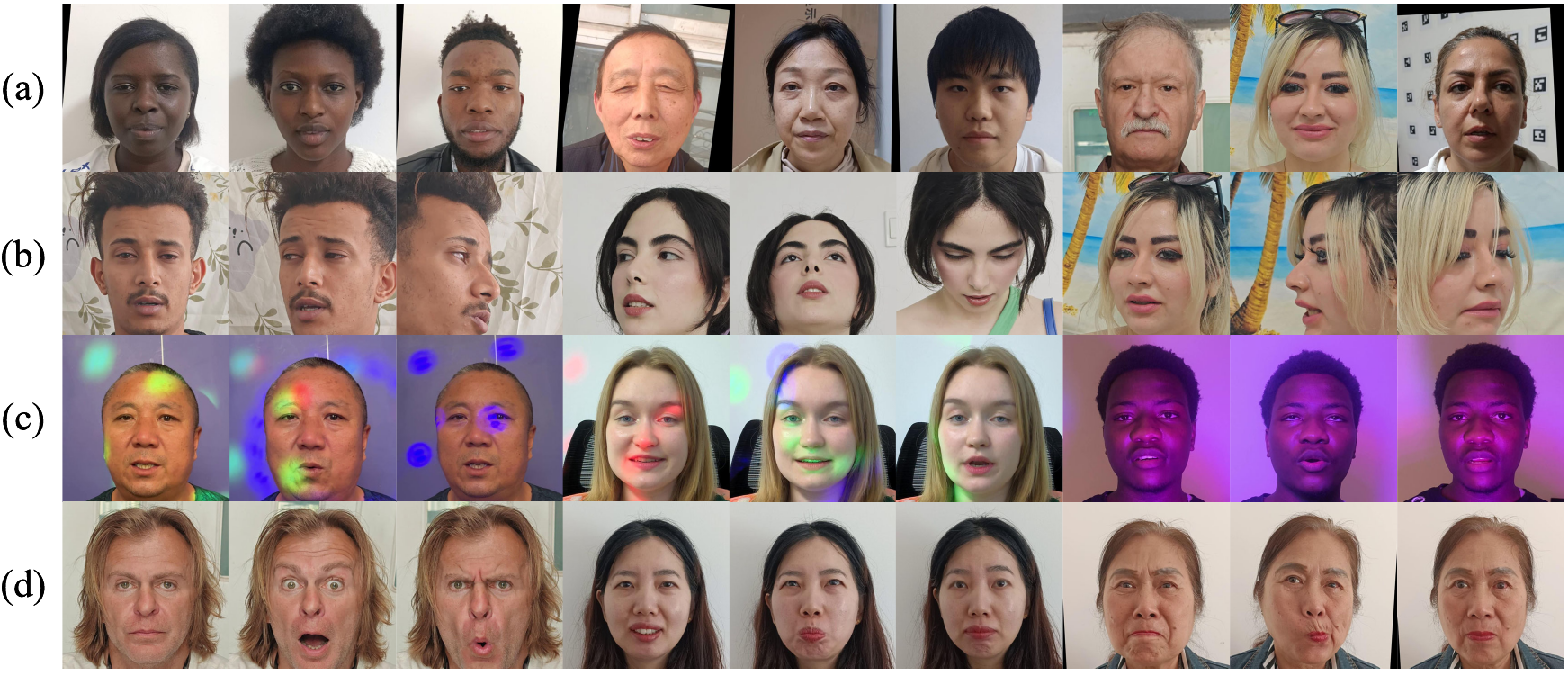}
  \end{center}
  \vspace{-0.2cm}
  \caption{Some of the aligned face images of the CASIA FaceSwapping database. From top to bottom, the rows show variations of the (a) ethnic,  (b) pose,  (c) illumination and (d) expression.
  \vspace{-0.2cm}
  }
\label{fig:database-examples}
\end{figure}

The CASIA FaceSwapping database is a large scale face database designed to evaluate different face swapping algorithms, which is recorded  using the mobile phone. Both the video and audio have been recorded. It includes 1,291 individuals and 2,582 videos, with variations in pose, expression,  and illumination. Details on data acquisition, data statics, and face preprocessing are provided below. The total frames are 2,826,644.
To imitate realistic scenarios, the subject  sits in front of the camera and speaks randomly during the recording.
The dataset comprises subjects with a roughly balanced racial distribution: 447 Asians, 372 Africans, and 472 Caucasians. Each subject contributed two video recordings, one with minimal variation which we denoted as normal and the other exhibiting more dynamic changes. The latter includes variations in pose, expression, or illumination, with each subject randomly assigned one type of variation.
In total, the dataset contains 1,291 static videos, 394 videos with pose variations, 564 with illumination changes, and 333 with expression changes.
Notably, all raw videos are captured in 4K resolution at 2160 × 3840, which distinguishes our dataset from others that often rely on lower-resolution recordings.

Given the substantial storage requirements of raw videos, we also generate output videos at different quality levels to better reflect realistic online content. Since uncompressed videos are rarely encountered on the internet, we apply H.265 compression, which is commonly used by social media and video-sharing platforms. High-quality videos are generated with light compression (quantization parameter set to 18), resulting in visually near-lossless output. In contrast, low-quality videos are compressed with a quantization parameter of 28. All videos are encoded at a frame rate of 30 \emph{fps}. 
To construct a consistent set of face images, we first uniformly sample 10 frames from each video sequence, yielding 25,820 frames in total. Face regions are then detected and aligned using the multitask cascaded convolutional neural network (MTCNN)~\cite{zhang2016joint}. Each frame is manually inspected, and if no face or an incorrect face is detected, it is replaced by a nearby frame with a valid detection. This results in a final set of 25,820  aligned face images. 
Some of the aligned face images are shown in Figure~\ref{fig:database-examples}. From top to bottom, the rows show variations of the ethnic, pose, illumination and expression.

\subsection{Evaluation Protocols}

\begin{figure}[t]
  \begin{center}
  \includegraphics[width=1.0\linewidth]{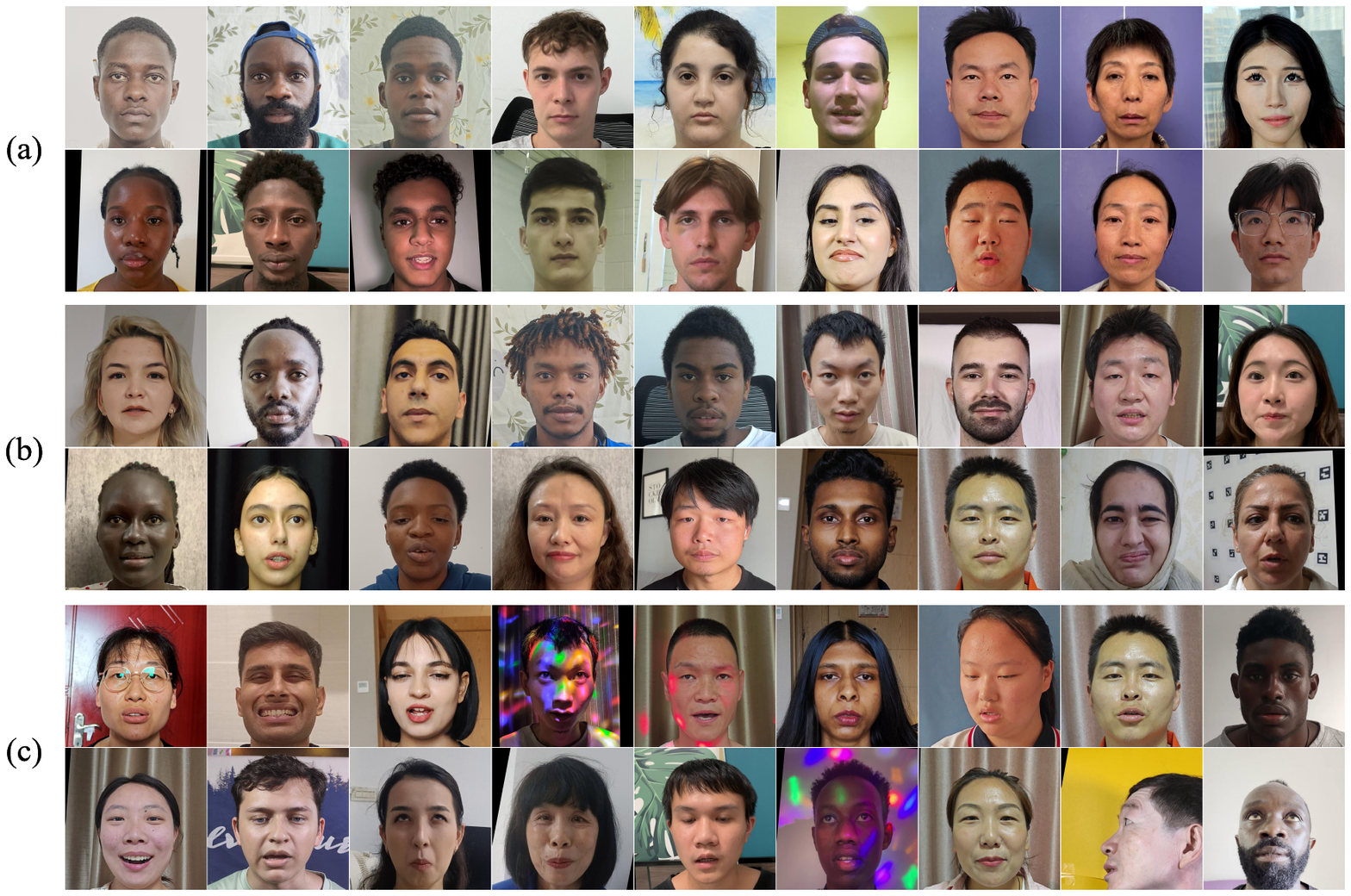}
  \end{center}
  \vspace{-0.2cm}
  \caption{Samples of different protocols: (a) Normal protocol,  (b) Cross-ethnicity protocol,  (c) Cross-attribute protocol. For each of the protocol, the first row represents the source images, the second row denotes the target images. 
  \vspace{-0.2cm}
  }
\label{fig:protocol-examples}
\end{figure}

To enable a fair, comprehensive, and fine-grained evaluation of face swapping algorithms, we establish the CASIA FaceSwapping benchmark with three standardized testing protocols. These protocols respectively target: (1) baseline performance under normal conditions, (2) generalization across cross-ethnicity settings, and (3) robustness to dynamic attribute variations such as pose, expression, and illumination. Each protocol is designed to isolate specific influencing factors and provides a well-structured setup for targeted and interpretable analysis. We have also shown the samples  of diffrent protocols in Figure~\ref{fig:protocol-examples}.
In the following, we present the construction principles and detailed configurations of the three protocols.

\vspace{1mm} \noindent \textbf{Protocol 1: Normal Protocol.} 
This protocol evaluates baseline performance under standard conditions. 
All video pairs are selected from normal recordings within the same ethnic group, removing demographic and attribute variations. Specifically, we construct 4,500 non-overlapping intra-ethnicity pairs, with about 1,500 pairs for each group: Asian, African, and Caucasian. This setting focuses on identity transfer and attribute preservation, serving as a baseline evaluation scenario.

\vspace{1mm} \noindent \textbf{Protocol 2: Cross-ethnicity Protocol.}
This protocol evaluates the generalization ability of face swapping algorithms when the source and target subjects belong to different ethnic groups.
By constructing cross-ethnicity video pairs while keeping all other factors consistent, this protocol reveals potential demographic biases and examines whether the algorithm can maintain identity transfer quality across diverse racial groups. It provides an essential fairness-oriented perspective that is missing in prior datasets.
Specifically, we generate 200 video pairs for each of the following cross-ethnicity combinations: Asian $\rightarrow$ African, Asian $\rightarrow$  Caucasian, and African $\rightarrow$  Caucasian. The corresponding reversed pairs (e.g., African  $\rightarrow$  Asian, Caucasian  $\rightarrow$  Asian, etc.) are also included by swapping the source and target roles, resulting in a total of 1,200 cross-ethnicity pairs.

\vspace{1mm} \noindent \textbf{Protocol 3: Cross-attribute Protocol.} This protocol evaluates the robustness of face swapping algorithms under dynamic changes in pose, expression, and illumination. This setting enables fine-grained analysis of how well algorithms preserve facial attributes and remain stable under attribute variations. Specifically, we sample 394 pose-varied videos, 333 expression-varied videos, and 564 illumination-varied videos. For each variation, we construct video pairs by matching normal videos with the corresponding varied videos. In total, we construct 4,300 non-overlapping video pairs across six categories: normal $\rightarrow$ pose, normal $\rightarrow$ illumination, normal $\rightarrow$ expression, and their corresponding reversed pairs.

\begin{table}[t]
\centering
\scriptsize
\caption{Summary of the three evaluation protocols in CASIA FaceSwapping, covering normal, cross-ethnicity, and cross-attribute settings.}
\label{tab:protocols}
\begin{tabular}{lccc}
\toprule
\textbf{Protocol} & \textbf{Ethnicity} & \textbf{Attribute} & \textbf{Pair Count} \\
\midrule
Protocol 1 (Normal)            & Same      & Normal                & 4,500 \\
\rowcolor{lightcyan} Protocol 2 (Cross-ethnicity)   & Different & Normal                & 1,200 \\
Protocol 3 (Cross-attribute)   & Same      & Different attributes  & 4,300 \\
\bottomrule
\end{tabular}
\end{table}

Compared with previous face swapping datasets, the proposed CASIA FaceSwapping dataset significantly improves in terms of resolution, subject diversity and demographic balance. 
Besides, it provides a more rigorous and fairness-aware benchmark tailored to the specific demands of face swapping evaluation. It enables fine-grained and ethically responsible analysis, which is crucial for advancing face manipulation technologies in real-world applications.

\begin{table}[t]
\centering
%\scriptsize
\footnotesize 
\renewcommand{\arraystretch}{1.25}
\setlength{\tabcolsep}{5pt}
\caption{Representative face swapping methods.}
\label{tab:reprentative_methods}
\begin{NiceTabular}{llll}
\CodeBefore
  \rowcolor{lightcyan}{2}
  \rowcolor{lightcyan}{10-13}
\Body
\toprule
\textbf{Category} & \textbf{Method} & \textbf{Publication} & \textbf{Code} \\
\midrule
\Block{1-1}{3D Model}
& HifiFace~\cite{hififace}
& IJCAI 2021
& \href{https://github.com/maum-ai/hififace}{Code} \\
\midrule
\Block{7-1}{GAN}
& FSGAN~\cite{nirkin2019fsgan,nirkin2022fsganv2}
& \makecell[l]{ICCV 2019 \\ TPAMI 2022}
& \href{https://github.com/YuvalNirkin/fsgan}{Code} \\
& FaceShifter~\cite{li2020advancing}
& CVPR 2020
& \href{https://github.com/maum-ai/faceshifter}{Code} \\
& BlendFace~\cite{shiohara2023blendface}
& ICCV 2023
& \href{https://github.com/mapooon/BlendFace}{Code} \\
& FaceDancer~\cite{rosberg2023facedancer}
& WACV 2023
& \href{https://github.com/felixrosberg/FaceDancer}{Code} \\
& SimSwap~\cite{chen2020simswap,chen2024simswap++}
& \makecell[l]{MM 2020 \\ TPAMI 2024}
& \href{https://github.com/neuralchen/SimSwap}{Code} \\
& CSCS~\cite{huang2024identity}
& TOG 2024
& \href{https://github.com/ICTMCG/CSCS}{Code} \\
& InsightFace~\cite{InsightFace-url}
& --
& \href{https://github.com/deepinsight/insightface}{Code} \\
\midrule
\Block{4-1}{StyleGAN}
& MegaFS~\cite{zhu2021one}
& CVPR 2021
& \href{https://github.com/zyainfal/One-Shot-Face-Swapping-on-Megapixels}{Code} \\
& FSLSD~\cite{xu2022high}
& CVPR 2022
& \href{https://github.com/cnnlstm/FSLSD_HiRes}{Code} \\
& RAFSwap~\cite{xu2022region}
& CVPR 2022
& \href{https://github.com/xc-csc101/RAFSwap}{Code} \\
& RGISwap~\cite{liu2023fine}
& CVPR 2023
& \href{https://github.com/e4s2022/e4s}{Code} \\
\midrule
\Block{2-1}{Diffusion}
& DiffSwap~\cite{zhao2023diffswap}
& CVPR 2023
& \href{https://github.com/wl-zhao/DiffSwap}{Code} \\
& Face-Adapter~\cite{han2024face}
& ECCV 2024
& \href{https://github.com/FaceAdapter/Face-Adapter}{Code} \\
\bottomrule
\end{NiceTabular}
\end{table}

\subsection{Evaluation Results}
Due to the diversity of face swapping methods, we benchmark representative open-source approaches on our dataset. 
The selected methods are listed in Table~\ref{tab:reprentative_methods}. 
For fairness, we use publicly available pre-trained models.

\begin{table*}[!t]
\centering
\small
\caption{Benchmark results across 14 face swapping methods evaluated on the proposed protocols. Identity preservation is measured by ID retrieval and ID similarity, while pose error, expression error, and FID  reflect the  attribute preservation and generation quality.}
\label{tab:main_results_multicolumn}
\begin{tabular}{llccccc}
\toprule
\textbf{\multirow{2}{*}{Method}} & \textbf{\multirow{2}{*}{Protocol}} &
\textbf{\multirow{2}{*}{\begin{tabular}[c]{@{}c@{}}ID retrieval $\uparrow$  \end{tabular}}} &
\textbf{\multirow{2}{*}{\begin{tabular}[c]{@{}c@{}}ID similarity $\uparrow$ \end{tabular}}} &
\textbf{\multirow{2}{*}{\begin{tabular}[c]{@{}c@{}}pose error $\downarrow$  \end{tabular}}} &
\textbf{\multirow{2}{*}{\begin{tabular}[c]{@{}c@{}}expression error $\downarrow$ \end{tabular}}} &
\textbf{\multirow{2}{*}{\begin{tabular}[c]{@{}c@{}}FID $\downarrow$  \end{tabular}}}  \\ 
  & & & & & & \\
\midrule
\cellcolor{lightcyan}
& \cellcolor{lightcyan}Normal 
& \cellcolor{lightcyan}93.37\% 
& \cellcolor{lightcyan}0.62 
& \cellcolor{lightcyan}3.59 
& \cellcolor{lightcyan}3.12 
& \cellcolor{lightcyan}20.40 \\
\cellcolor{lightcyan}HifiFace~\cite{hififace}
& \cellcolor{lightcyan}Cross-ethnicity 
& \cellcolor{lightcyan}93.23\% 
& \cellcolor{lightcyan}0.60 
& \cellcolor{lightcyan}3.66 
& \cellcolor{lightcyan}3.29 
& \cellcolor{lightcyan}21.73  \\
\cellcolor{lightcyan}
& \cellcolor{lightcyan}Cross-attribute 
& \cellcolor{lightcyan}83.57\% 
& \cellcolor{lightcyan}0.57 
& \cellcolor{lightcyan}4.12 
& \cellcolor{lightcyan}3.14 
& \cellcolor{lightcyan}9.93 \\
\midrule
\multirow{3}{*}{FSGAN~\cite{nirkin2019fsgan}} 
& Normal & 65.08\% & 0.50 & 3.34 & 2.35 & 56.23  \\
& Cross-ethnicity & 57.82\% & 0.44 & 3.44 & 2.50 & 58.06  \\
& Cross-attribute & 43.74\% & 0.40 & 4.08 & 2.37 & 40.83 \\
\midrule
\cellcolor{lightcyan}
& \cellcolor{lightcyan}Normal 
& \cellcolor{lightcyan}66.41\% 
& \cellcolor{lightcyan}0.44 
& \cellcolor{lightcyan}5.26 
& \cellcolor{lightcyan}3.64 
& \cellcolor{lightcyan}169.11 \\
\cellcolor{lightcyan}Faceshifter~\cite{li2020advancing}
& \cellcolor{lightcyan}Cross-ethnicity 
& \cellcolor{lightcyan}66.36\% 
& \cellcolor{lightcyan}0.43 
& \cellcolor{lightcyan}5.32 
& \cellcolor{lightcyan}3.78 
& \cellcolor{lightcyan}172.31 \\
\cellcolor{lightcyan}
& \cellcolor{lightcyan}Cross-attribute 
& \cellcolor{lightcyan}56.09\% 
& \cellcolor{lightcyan}0.40 
& \cellcolor{lightcyan}6.38 
& \cellcolor{lightcyan}3.67 
& \cellcolor{lightcyan}151.61 \\
\midrule
\multirow{3}{*}{BlendFace~\cite{shiohara2023blendface}} 
& Normal & 73.35\% & 0.48 & 3.28 & 3.08 & 93.20 \\
& Cross-ethnicity & 70.60\% & 0.45 & 3.37 & 3.19 & 94.51 \\
& Cross-attribute & 64.83\% & 0.44 & 3.93 & 3.07 & 78.67 \\
\midrule
\cellcolor{lightcyan}
& \cellcolor{lightcyan}Normal 
& \cellcolor{lightcyan}72.81\% 
& \cellcolor{lightcyan}0.49 
& \cellcolor{lightcyan}3.42 
& \cellcolor{lightcyan}3.15 
& \cellcolor{lightcyan}19.14 \\
\cellcolor{lightcyan}FaceDancer~\cite{rosberg2023facedancer}
& \cellcolor{lightcyan}Cross-ethnicity 
& \cellcolor{lightcyan}78.74\% 
& \cellcolor{lightcyan}0.50 
& \cellcolor{lightcyan}3.72 
& \cellcolor{lightcyan}3.56 
& \cellcolor{lightcyan}22.33 \\
\cellcolor{lightcyan}
& \cellcolor{lightcyan}Cross-attribute 
& \cellcolor{lightcyan}62.49\% 
& \cellcolor{lightcyan}0.46 
& \cellcolor{lightcyan}3.95 
& \cellcolor{lightcyan}3.14 
& \cellcolor{lightcyan}6.32 \\
\midrule
\multirow{3}{*}{SimSwap~\cite{chen2020simswap}} 
& Normal & 90.00\% & 0.61 & 2.14 & 2.43 & 21.75  \\
& Cross-ethnicity & 90.74\% & 0.58 & 2.21 & 2.63 & 24.01  \\
& Cross-attribute & 81.50\% & 0.55 & 2.43 & 2.42 & 7.86  \\
\midrule
\cellcolor{lightcyan}
& \cellcolor{lightcyan}Normal 
& \cellcolor{lightcyan}88.75\% 
& \cellcolor{lightcyan}0.63 
& \cellcolor{lightcyan}3.81 
& \cellcolor{lightcyan}3.41 
& \cellcolor{lightcyan}33.28  \\
\cellcolor{lightcyan}CSCS~\cite{huang2024identity}
& \cellcolor{lightcyan}Cross-ethnicity 
& \cellcolor{lightcyan}96.92\% 
& \cellcolor{lightcyan}0.65 
& \cellcolor{lightcyan}4.11 
& \cellcolor{lightcyan}3.72 
& \cellcolor{lightcyan}36.17  \\
\cellcolor{lightcyan}
& \cellcolor{lightcyan}Cross-attribute 
& \cellcolor{lightcyan}87.54\% 
& \cellcolor{lightcyan}0.60 
& \cellcolor{lightcyan}4.47 
& \cellcolor{lightcyan}3.44 
& \cellcolor{lightcyan}21.23 \\
\midrule
\multirow{3}{*}{InsightFace~\cite{InsightFace-url}}
& Normal & 96.92\% & 0.73 & 2.84 & 2.64 & 30.50 \\
& Cross-ethnicity & 97.19\% & 0.71 & 2.97 & 2.87 & 32.32  \\
& Cross-attribute & 95.14\% & 0.67 & 3.22 & 2.62 & 15.86 \\
\midrule
\cellcolor{lightcyan}
& \cellcolor{lightcyan}Normal 
& \cellcolor{lightcyan}73.70\% 
& \cellcolor{lightcyan}0.50 
& \cellcolor{lightcyan}5.09 
& \cellcolor{lightcyan}2.96 
& \cellcolor{lightcyan}23.69 \\
\cellcolor{lightcyan}MegaFS~\cite{zhu2021one}
& \cellcolor{lightcyan}Cross-ethnicity 
& \cellcolor{lightcyan}72.72\% 
& \cellcolor{lightcyan}0.49 
& \cellcolor{lightcyan}5.09 
& \cellcolor{lightcyan}3.15 
& \cellcolor{lightcyan}25.93 \\
\cellcolor{lightcyan}
& \cellcolor{lightcyan}Cross-attribute 
& \cellcolor{lightcyan}55.82\% 
& \cellcolor{lightcyan}0.44 
& \cellcolor{lightcyan}5.98 
& \cellcolor{lightcyan}3.02 
& \cellcolor{lightcyan}18.24  \\
\midrule
\multirow{3}{*}{FSLSD~\cite{xu2022high}}
& Normal & 15.52\% & 0.25 & 5.63 & 3.44 & 28.64\\
& Cross-ethnicity & 13.95\% & 0.23 & 5.62 & 3.58 & 30.47 \\
& Cross-attribute & 11.95\% & 0.23 & 7.24 & 3.54 & 23.24 \\
\midrule
\cellcolor{lightcyan}
& \cellcolor{lightcyan}Normal 
& \cellcolor{lightcyan}87.77\% 
& \cellcolor{lightcyan}0.54 
& \cellcolor{lightcyan}3.69 
& \cellcolor{lightcyan}3.28 
& \cellcolor{lightcyan}45.61 \\
\cellcolor{lightcyan}RAFSwap~\cite{xu2022region}
& \cellcolor{lightcyan}Cross-ethnicity 
& \cellcolor{lightcyan}86.00\% 
& \cellcolor{lightcyan}0.51 
& \cellcolor{lightcyan}3.74 
& \cellcolor{lightcyan}3.46 
& \cellcolor{lightcyan}47.47 \\
\cellcolor{lightcyan}
& \cellcolor{lightcyan}Cross-attribute 
& \cellcolor{lightcyan}72.40\% 
& \cellcolor{lightcyan}0.48 
& \cellcolor{lightcyan}4.80 
& \cellcolor{lightcyan}3.31 
& \cellcolor{lightcyan}31.37 \\
\midrule
\multirow{3}{*}{RGISwap~\cite{liu2023fine}}
& Normal & 80.84\% & 0.53 & 4.00 & 3.41 & 18.77 \\
& Cross-ethnicity & 80.92\% & 0.52 & 4.03 & 3.58 & 21.28 \\
& Cross-attribute & 62.96\% & 0.46 & 4.90 & 3.54 & 13.94 \\
\midrule
\cellcolor{lightcyan}
& \cellcolor{lightcyan}Normal 
& \cellcolor{lightcyan}15.64\% 
& \cellcolor{lightcyan}0.32 
& \cellcolor{lightcyan}3.67 
& \cellcolor{lightcyan}2.88 
& \cellcolor{lightcyan}96.55  \\
\cellcolor{lightcyan}DiffSwap~\cite{zhao2023diffswap}
& \cellcolor{lightcyan}Cross-ethnicity 
& \cellcolor{lightcyan}13.70\% 
& \cellcolor{lightcyan}0.27 
& \cellcolor{lightcyan}3.74 
& \cellcolor{lightcyan}2.97 
& \cellcolor{lightcyan}97.52 \\
\cellcolor{lightcyan}
& \cellcolor{lightcyan}Cross-attribute 
& \cellcolor{lightcyan}14.38\% 
& \cellcolor{lightcyan}0.30 
& \cellcolor{lightcyan}4.14 
& \cellcolor{lightcyan}2.89 
& \cellcolor{lightcyan}87.11 \\
\midrule
\multirow{3}{*}{FaceAdapter~\cite{han2024face}}
& Normal & 95.49\% & 0.66 & 4.38 & 2.95 & 23.83  \\
& Cross-ethnicity & 94.74\% & 0.66 & 4.83 & 3.22 & 26.51  \\
& Cross-attribute & 88.81\% & 0.61 & 5.05 & 2.96 & 14.71 \\
\bottomrule
\end{tabular}
\end{table*}

\begin{figure*}[t]
  \begin{center}
  \includegraphics[width=1.0\linewidth]{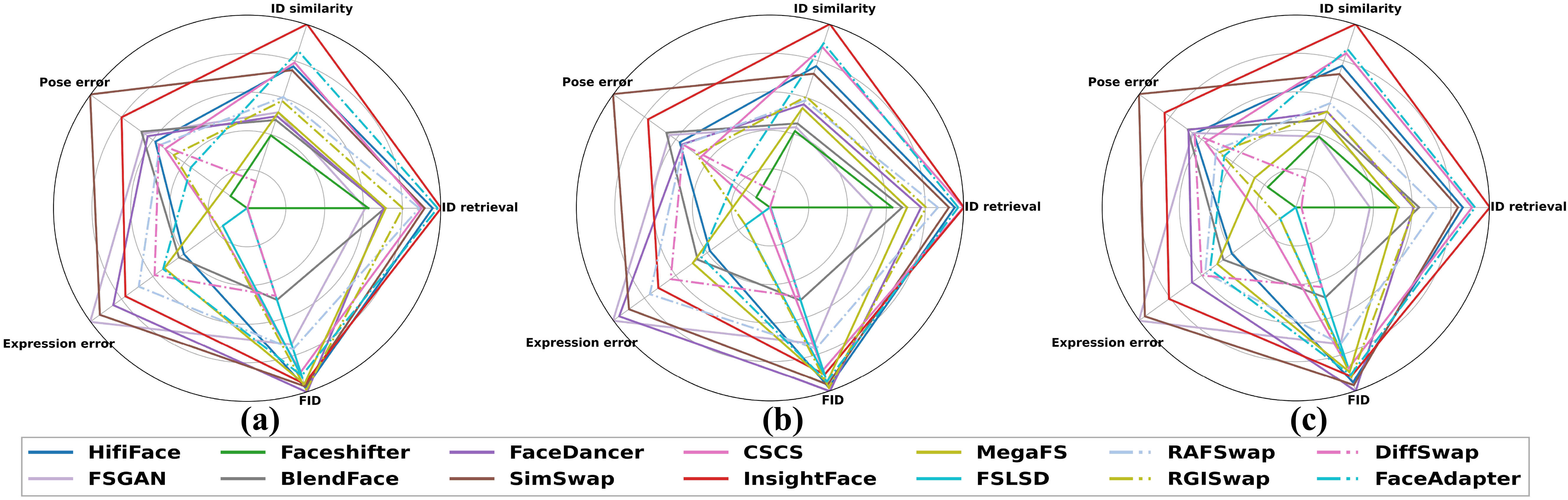}
  \end{center}
  \vspace{-0.2cm}
 \caption{Normalized radar charts of face swapping methods across the proposed protocols: (a) Normal, (b) Cross-ethnicity, and (c) Cross-attribute. Each chart summarizes five metrics: ID retrieval, ID similarity, pose error, expression error, and FID, reflecting identity preservation, attribute consistency, and visual fidelity. Values are min-max normalized within each protocol, and pose error, expression error, and FID are inverted for visualization. 
\vspace{-0.2cm}
}
\label{fig:radar-chart}
\end{figure*}

\begin{figure*}[t]
  \begin{center}
  \includegraphics[width=1.0\linewidth]{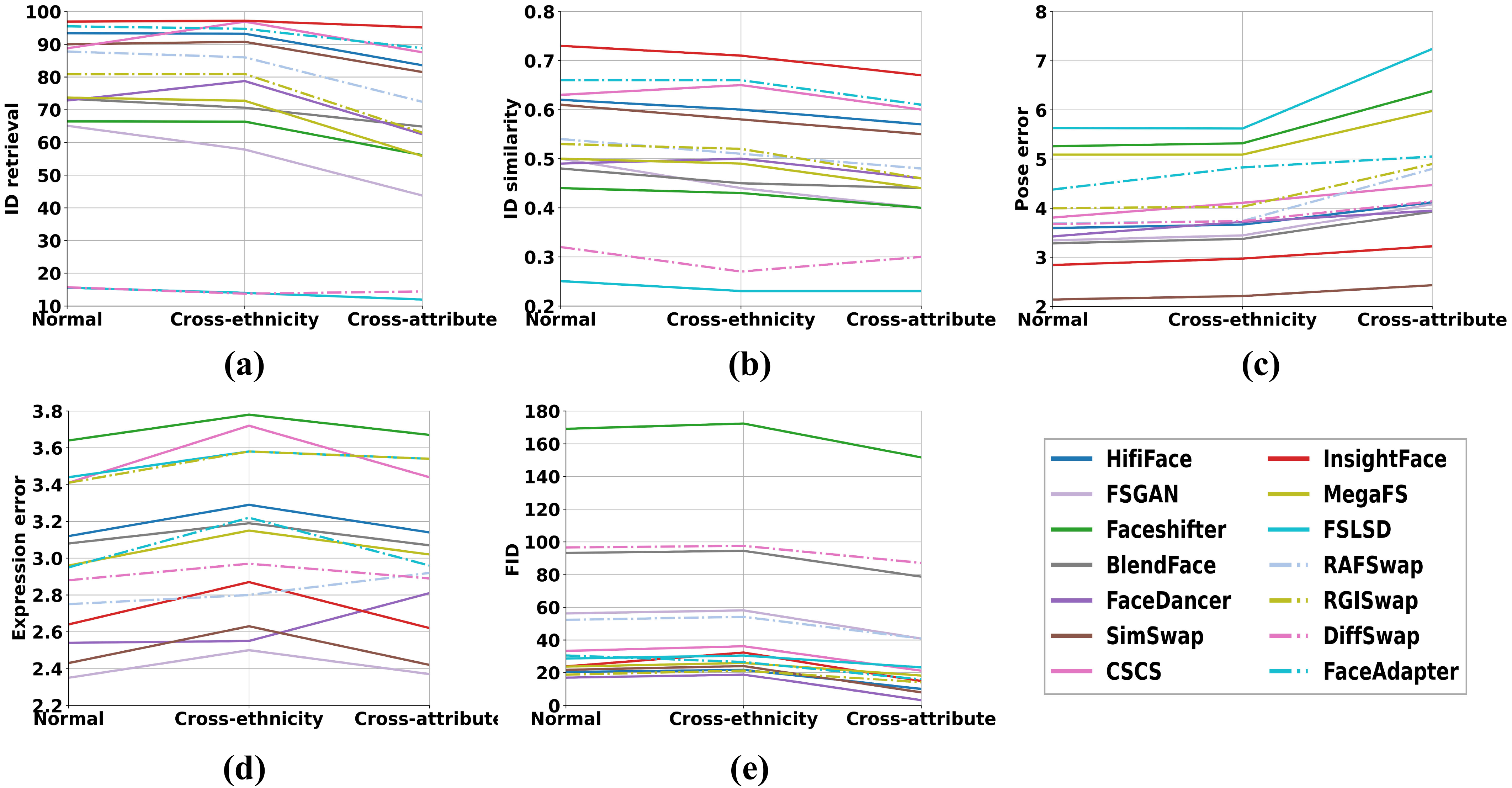}
  \end{center}
  \vspace{-0.2cm}
  \caption{Line chart of face swapping methods across the proposed protocols for different metrics: (a) ID retrieval, (b) ID similarity, (c) pose error, (d) expression error, and (e) FID. Each subplot corresponds to a single metric, showing the performance trend of all methods under different protocols. 
  \vspace{-0.2cm}
  }
\label{fig:trend-image}
\end{figure*}

\vspace{1mm} \noindent \textbf{Quantitative Comparison under Diverse Protocols.}
Table~\ref{tab:main_results_multicolumn} presents a comprehensive evaluation of face swapping methods under three protocols: Normal, Cross-ethnicity, and Cross-attribute, using six metrics (Section~\ref{subsec:evaluation-metrics}). These protocols assess robustness to demographic and attribute variations.

ID retrieval and ID similarity reflect identity preservation. InsightFace achieves the strongest performance across all protocols, with ID retrieval of 96.92\%, 97.19\%, and 95.14\%, and ID similarity above 0.67.  This makes it the top performer in both recognition-driven metrics.  FaceAdapter follows, maintaining ID retrieval above 88\% and ID similarity above 0.66, indicating robust identity consistency even under cross-ethnicity and cross-attribute settings. HifiFace ranks next, performing well under Normal and Cross-ethnicity but slightly degrading under Cross-attribute. In contrast, FSLSD and DiffSwap perform poorly, with ID retrieval below 16\% and ID similarity under 0.35, indicating severe identity collapse. These methods fail to maintain consistent identity traits, particularly when dealing with diverse attributes or demographics.

Pose and expression errors measure attribute preservation. SimSwap achieves the lowest pose errors (2.14, 2.21, 2.43). Its stable performance across all settings suggests effective disentanglement between identity and pose, enabling precise spatial control during face swapping. InsightFace also performs well,  showing architectural alignment between identity encoding and geometric consistency. For expression, FSGAN and InsightFace achieve errors below 2.9, preserving fine facial dynamics. FaceDancer shows moderate errors (3.14--3.56), comparable to BlendFace and CSCS, and better than FaceShifter, which shows the largest distortion (up to 3.78). These results suggest that aggressive warping or blending may degrade local semantics.

It is worth emphasizing that identity preservation and attribute consistency are intrinsically competing objectives in face swapping. In an extreme case, simply outputting the target face would completely discard the source identity, leading to near-zero ID retrieval, while achieving minimal pose and expression errors, as all target attributes are perfectly preserved. This observation highlights that strong attribute fidelity alone does not imply successful face swapping. Therefore, an effective evaluation protocol should explicitly balance identity retention and attribute preservation, favoring methods that maintain both aspects simultaneously rather than optimizing either in isolation.

FID measures visual quality. FaceDancer achieves low FID scores (19.14, 22.33, 6.32), indicating strong visual realism, though with weaker identity performance than InsightFace and SimSwap. SimSwap, HifiFace, and RGISwap also maintain FID below 25. Notably, FaceAdapter achieves a strong balance between realism and identity fidelity, making it a practical candidate for deployment. In contrast, FaceShifter (FID $>$150) and DiffSwap (FID $>$90) show severe visual degradation, often with artifacts and distortion under attribute variations.

Figure~\ref{fig:radar-chart} also presents a normalized radar chart summarizing the performance of 14 face swapping methods across different protocols. As shown in the figure, InsightFace and SimSwap present the most balanced profiles across all metrics. FaceDancer, while achieving one of the lowest FID scores, also demonstrates solid pose and expression consistency, though its identity metrics remain moderately lower than top-performing methods.
Overall, InsightFace emerges as the most well-rounded method across all evaluation dimensions. SimSwap exhibits strong robustness in pose and expression preservation, particularly under cross-domain conditions. FaceAdapter also achieves a competitive balance between identity preservation and visual realism, making it a strong candidate for practical deployment. In contrast, methods such as Faceshifter, FSLSD, and DiffSwap show clear limitations across multiple metrics, underscoring the need for improved architectures and training objectives that better disentangle identity from attributes while ensuring high-fidelity synthesis.

Figure~\ref{fig:trend-image} illustrates protocol-level performance trends. As expected, most methods experience a performance drop under Cross-ethnicity and Cross-attribute protocols compared to Normal protocol. For example, HifiFace declines from ID retrieval in Normal to 83.57\%  in Cross-attribute. However, CSCS shows an unusual trend: its ID retrieval improves from 88.75\%  (Normal) to 96.92\%  (Cross-ethnicity), possibly due to enhanced modeling of identity-conditioned generation across race boundaries.

\begin{figure*}[t]
  \begin{center}
  \includegraphics[width=1.0\linewidth]{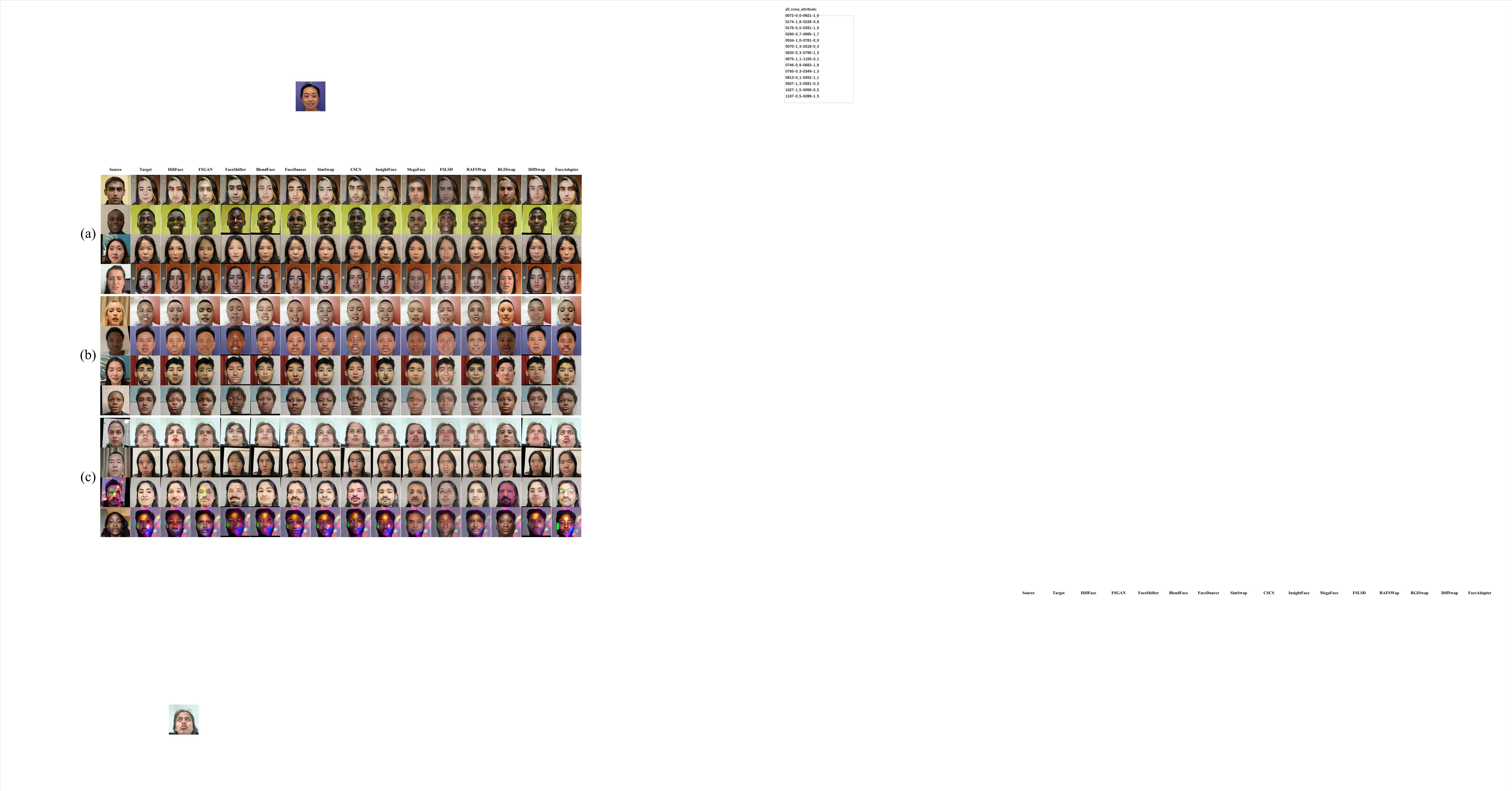}
  \end{center}
  \vspace{-0.2cm}
  \caption{Qualitative results of face swapping methods across different  protocols. (a) Normal protocol (top four rows), (b) Cross-ethnicity protocol (middle four rows), and (c) Cross-attribute protocol (bottom four rows). Best viewed by zooming in.
  \vspace{-0.2cm}
  }
\label{fig:total-protocol-qualitative-comparison}
\end{figure*}

\vspace{1mm} \noindent \textbf{Qualitative Comparison under Diverse Protocols.}
Figure~\ref{fig:total-protocol-qualitative-comparison} provides qualitative comparisons complementing the quantitative results. Under the Normal protocol (Figure~\ref{fig:total-protocol-qualitative-comparison}(a)), most methods generate plausible results, but differences remain in preserving fine-grained identity cues and suppressing minor artifacts, especially near facial boundaries. Methods such as FSLSD and DiffSwap exhibit artifacts or over-smoothing, consistent with Table~\ref{tab:main_results_multicolumn} and Figure~\ref{fig:radar-chart}.

Under the Cross-ethnicity protocol (Figure~\ref{fig:total-protocol-qualitative-comparison}(b)), differences become more pronounced. Identity leakage appears in some methods, reflected in shifts in facial geometry and appearance (e.g., DiffSwap and MegaFS). Skin-tone inconsistencies and boundary artifacts are also observed (e.g., FSGAN and FaceShifter). In contrast, InsightFace and FaceAdapter better preserve identity while adapting appearance, aligning with stronger quantitative performance.

The Cross-attribute protocol (Figure~\ref{fig:total-protocol-qualitative-comparison}(c)) is the most challenging, with large variations in pose, expression, and illumination. Failures often occur in deformation-sensitive regions (e.g., mouth), leading to inconsistent expressions or distorted shapes. Some methods show local warping and boundary seams (e.g., CSCS and FSLSD), indicating weak disentanglement. Methods such as InsightFace, SimSwap, and FaceAdapter produce more coherent results, consistent with their smaller performance degradation.

Overall, qualitative differences are most evident under Cross-ethnicity and Cross-attribute protocols, where identity and attribute consistency are jointly stressed. Together with quantitative results, these observations suggest that failures stem from imperfect disentanglement and unstable local reconstruction under distribution shifts.

\vspace{1mm} \noindent \textbf{Temporal Consistency.} 
We evaluate temporal consistency on 30 videos, each using a 100-frame segment. Methods are tested at $2160\times3840$ on full videos and $256\times256$ on cropped faces. Following Section~\ref{subsec:evaluation-metrics}, we use subject and background consistency, computed as average cosine similarity across frames relative to the first and preceding frames, to measure identity and background stability. Results are summarized in Table~\ref{tab:vbench_metrics}, enabling comparison of temporal stability.

\begin{table*}[t]
\centering
\small
\caption{
Temporal consistency comparison of face swapping methods on video clips. Subject consistency and background consistency are computed to measure frame-to-frame stability of the swapped face region and the surrounding background under three protocols at $2160\times3840$ and $256\times256$ resolutions. The final column reports the overall average of the subject consistency and background consistency scores.
}
\label{tab:vbench_metrics}
\resizebox{\textwidth}{!}{
\begin{NiceTabular}{@{}llccccccc@{}}
\CodeBefore
  \rowcolor{lightcyan}{3,4}
  \rowcolor{lightcyan}{7,8}
  \rowcolor{lightcyan}{11,12}
  \rowcolor{lightcyan}{15,16}
  \rowcolor{lightcyan}{19,20}
  \rowcolor{lightcyan}{23,24}
  \rowcolor{lightcyan}{27,28}
\Body
\toprule
\Block{2-1}{\textbf{Method}} & \Block{2-1}{\textbf{Resolution}} &
\Block{1-3}{\textbf{Subject consistency (\%)} $\uparrow$} &
&
&
\Block{1-3}{\textbf{Background consistency (\%)} $\uparrow$} &
&
&
\Block{2-1}{\textbf{Average} $\uparrow$} \\
\cmidrule(lr){3-5} \cmidrule(lr){6-8}
& & \textbf{Normal} & \textbf{Cross-ethnicity} & \textbf{Cross-attribute}
& \textbf{Normal} & \textbf{Cross-ethnicity} & \textbf{Cross-attribute} & \\
\midrule

\Block{2-1}{HifiFace~\cite{hififace}}
& $2160\times3840$ & 99.33 & 99.29 & 98.99 & 97.67 & 97.60 & 97.56 & 98.41 \\
& $256\times256$   & 97.74 & 98.44 & 97.24 & 97.18 & 97.30 & 96.45 & 97.40 \\
\midrule

\Block{2-1}{FSGAN~\cite{nirkin2019fsgan}}
& $2160\times3840$ & 99.25 & 99.27 & 98.86 & 97.35 & 97.53 & 97.00 & 98.21 \\
& $256\times256$   & 98.28 & 98.25 & 97.83 & 96.64 & 96.38 & 96.05 & 97.24 \\
\midrule

\Block{2-1}{Faceshifter~\cite{li2020advancing}}
& $2160\times3840$ & 99.24 & 99.25 & 98.87 & 97.84 & 97.91 & 97.53 & 98.44 \\
& $256\times256$   & 97.78 & 98.42 & 97.11 & 97.51 & 97.85 & 97.23 & 97.65 \\
\midrule

\Block{2-1}{BlendFace~\cite{shiohara2023blendface}}
& $2160\times3840$ & 99.34 & 99.31 & 98.98 & 97.16 & 97.57 & 96.98 & 98.23 \\
& $256\times256$   & 97.72 & 98.44 & 97.08 & 96.30 & 97.20 & 96.31 & 97.18 \\
\midrule

\Block{2-1}{FaceDancer~\cite{rosberg2023facedancer}}
& $2160\times3840$ & 99.33 & 99.30 & 98.93 & 97.87 & 98.02 & 97.27 & 98.46 \\
& $256\times256$   & 97.69 & 98.34 & 96.96 & 97.16 & 97.69 & 96.53 & 97.40 \\
\midrule

\Block{2-1}{SimSwap~\cite{chen2020simswap}}
& $2160\times3840$ & 99.38 & 99.25 & 98.95 & 97.46 & 96.98 & 96.91 & 98.16 \\
& $256\times256$   & 97.77 & 98.41 & 97.14 & 96.64 & 97.46 & 96.39 & 97.30 \\
\midrule

\Block{2-1}{CSCS~\cite{huang2024identity}}
& $2160\times3840$ & 99.25 & 99.27 & 98.87 & 97.48 & 97.50 & 96.97 & 98.23 \\
& $256\times256$   & 97.62 & 98.33 & 97.10 & 97.01 & 97.21 & 96.77 & 97.34 \\
\midrule

\Block{2-1}{InsightFace~\cite{InsightFace-url}}
& $2160\times3840$ & 99.34 & 99.32 & 99.02 & 98.12 & 98.02 & 97.46 & 98.55 \\
& $256\times256$   & 97.59 & 97.67 & 96.64 & 97.43 & 97.50 & 96.91 & 97.29 \\
\midrule

\Block{2-1}{MegaFS~\cite{zhu2021one}}
& $2160\times3840$ & 99.06 & 99.13 & 98.79 & 97.25 & 97.46 & 96.86 & 98.09 \\
& $256\times256$   & 97.35 & 98.01 & 96.74 & 96.33 & 96.89 & 95.84 & 96.86 \\
\midrule

\Block{2-1}{FSLSD~\cite{xu2022high}}
& $2160\times3840$ & 98.70 & 98.90 & 97.86 & 96.76 & 97.13 & 94.85 & 97.37 \\
& $256\times256$   & 94.88 & 96.19 & 91.85 & 95.90 & 96.48 & 94.92 & 95.04 \\
\midrule

\Block{2-1}{RAFSwap~\cite{xu2022region}}
& $2160\times3840$ & 99.28 & 99.23 & 98.87 & 96.85 & 96.87 & 96.38 & 97.92 \\
& $256\times256$   & 97.81 & 98.32 & 97.05 & 95.90 & 96.85 & 95.52 & 96.91 \\
\midrule

\Block{2-1}{RGISwap~\cite{liu2023fine}}
& $2160\times3840$ & 99.25 & 99.22 & 98.86 & 97.33 & 97.35 & 96.95 & 98.16 \\
& $256\times256$   & 97.58 & 98.37 & 97.17 & 96.62 & 97.30 & 96.80 & 97.31 \\
\midrule

\Block{2-1}{DiffSwap~\cite{zhao2023diffswap}}
& $2160\times3840$ & 98.88 & 98.81 & 98.47 & 95.64 & 94.98 & 95.12 & 96.99 \\
& $256\times256$   & 95.58 & 96.36 & 94.89 & 94.86 & 94.78 & 94.50 & 95.16 \\
\midrule

\Block{2-1}{FaceAdapter~\cite{han2024face}}
& $2160\times3840$ & 99.18 & 99.11 & 98.78 & 97.73 & 97.56 & 97.14 & 98.26 \\
& $256\times256$   & 97.11 & 97.48 & 96.34 & 96.66 & 96.91 & 96.48 & 96.83 \\

\bottomrule
\end{NiceTabular}
}
\end{table*}

At $256\times256$, most methods show degraded temporal consistency. Some maintain strong subject consistency (e.g., FSGAN $\sim$98\%), while background consistency is more affected. FaceShifter performs best, achieving 97.65\% overall, with background consistency above 97\% across protocols, suggesting that background and structure dominate temporal quality at low resolution. In contrast, FSLSD shows a sharp drop in subject consistency (down to 91.85\% under Cross-attribute), indicating sensitivity to attribute shifts.

Across protocols, cross-domain settings degrade performance, but failure modes differ: some methods degrade in subject consistency, others in background consistency. At $256\times256$, many methods retain high subject consistency under Cross-ethnicity ($\geq$98\%), while Cross-attribute more often causes subject instability, reflecting the difficulty of disentanglement and local reconstruction. Background fluctuations indicate weaker constraints on non-face regions.

Overall, subject consistency often saturates, while background consistency and low-resolution robustness differentiate methods. InsightFace performs best at high resolution, while FaceShifter excels at low resolution. These results highlight the need for stronger temporal modeling, improved background consistency, and more robust identity--attribute disentanglement under challenging settings.

\section{Conclusion and Outlook}
\label{sec:conclusion}

In this paper, we presented a comprehensive study of face swapping from methodological and evaluative perspectives. Methodologically, we reviewed existing approaches under a unified taxonomy, covering five paradigms: 3DMM-based, autoencoder-based, GAN-based, StyleGAN-based, and diffusion-based methods, analyzing their representations, generation mechanisms, and identity--attribute disentanglement strategies. From the evaluation perspective, we introduced the CASIA FaceSwapping benchmark with standardized protocols and metrics. Extensive experiments on representative methods provide a systematic view of their strengths and limitations.

Face swapping has evolved from geometry- and reconstruction-based pipelines to expressive generative systems, improving visual fidelity and controllability. However, challenges remain in identity--attribute disentanglement, cross-domain generalization, and temporal consistency. We outline several promising directions for future research.

\vspace{1mm} \noindent \textbf{Building Facial Foundation Models.} Current methods remain prone to identity degradation under out-of-distribution data. A promising direction is large-scale facial foundation models trained with self-supervised objectives (e.g., masked image modeling), enabling more robust representations and improved generalization.

\vspace{1mm} \noindent \textbf{Integrating 3D Priors for Consistency.} Existing 2D methods often suffer from structural distortions under large pose or lighting changes. Incorporating 3D priors, via implicit supervision or explicit representations such as NeRF and 3D Gaussian Splatting, provides a principled path toward improved spatial consistency and multi-view coherence.

\vspace{1mm} \noindent \textbf{Pursuing Fine-grained Disentanglement.} Entanglement between identity and attributes remains a key challenge. Future work should enable finer control using explicit conditions (e.g., normal maps, identity-free textures, modular controllers), allowing independent manipulation of facial factors. Multimodal guidance (e.g., audio or language) may further enhance controllability.

\vspace{1mm} \noindent \textbf{Achieving Long-range Temporal Coherence.} Applying image-based models to videos often leads to flickering and identity drift. Future research should emphasize long-range temporal modeling, such as canonical-space representations and sequence-level consistency objectives, to improve stability over time.

\vspace{1mm} \noindent \textbf{Improving Efficiency for Real-world Deployment.} Many advanced methods, particularly diffusion-based ones, remain computationally expensive. Improving efficiency via distillation, few-step generation, and plug-and-play temporal refinement can enable practical deployment on resource-constrained devices.

\if 0
\ifCLASSOPTIONcompsoc
  \section*{Acknowledgments}
\else
  \section*{Acknowledgment}
\fi

%This work was supported in part by the National Key Research and Development Program of China under Grant 2022YFC3310400, in part by the National Natural Science Foundation of China (Grant Nos. U23B2054, 62276263,62102419), and the Youth Innovation Promotion Association CAS (Grant No. Y2023143).
\fi

\ifCLASSOPTIONcaptionsoff
  \newpage
\fi

\bibliographystyle{IEEEtran}
\bibliography{mytrans}

\vspace{-12mm}
\begin{IEEEbiography}[{\includegraphics[width=1in,height=1.25in,clip,keepaspectratio]{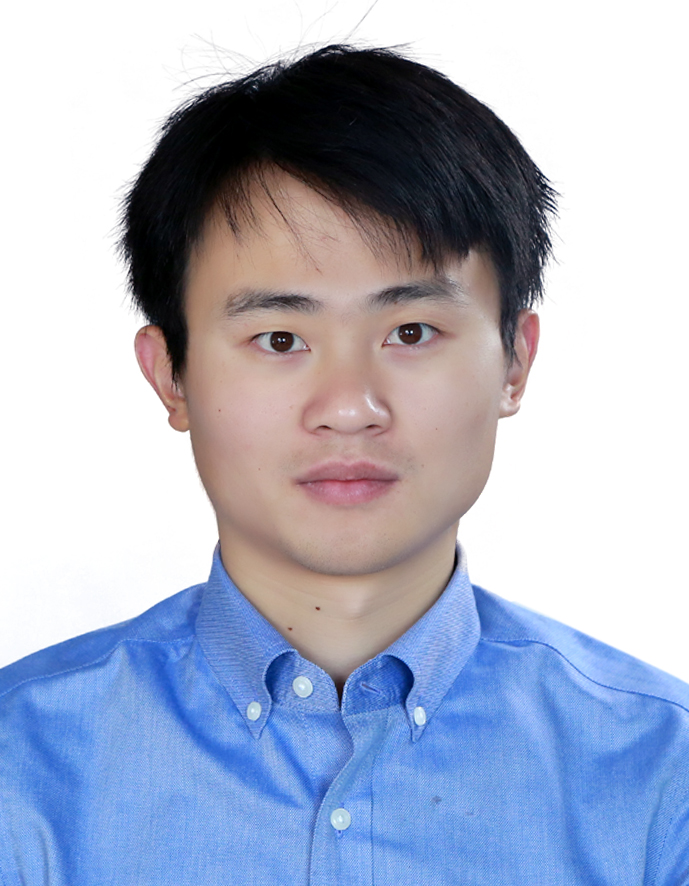}}]{Qi Li}
  received the B.E. degree from China University of Petroleum in 2011, the Ph.D. degree from the Institute of Automation, Chinese Academy of Sciences (CASIA) in 2016.
  He is an Associate Professor with the New Laboratory of Pattern Recognition (NLPR), State Key Laboratory of Multimodal Artificial Intelligence Systems (MAIS), CASIA. 
  His research interests include face recognition, computer vision, and machine learning.
\end{IEEEbiography}

\vspace{-12mm}
\begin{IEEEbiography}[{\includegraphics[width=1in,height=1.25in,clip,keepaspectratio]{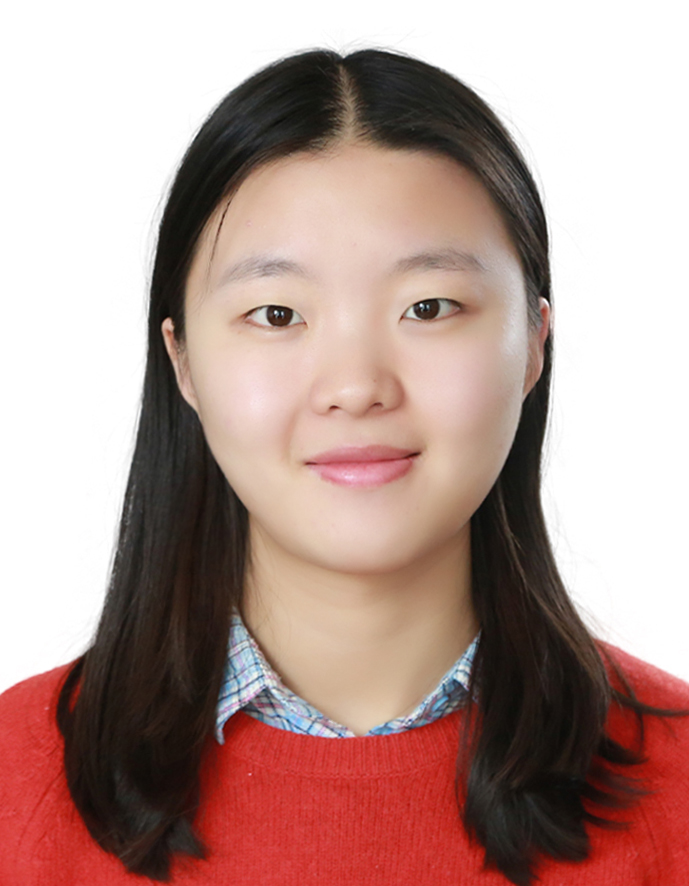}}]{Weining Wang}
  received her B.E. degree from North China Electric Power University in 2015 and the Ph.D. degree from University of Chinese Academy of Sciences (UCAS) in 2020. She is now an Associate Professor at the Laboratory of Cognition and Decision Intelligence for Complex Systems, Institute of Automation, Chinese Academy of Sciences (CASIA). Her research interests include pattern recognition and computer vision.
\end{IEEEbiography}

\vspace{-12mm}
\begin{IEEEbiography}[{\includegraphics[width=1in,height=1.25in,clip,keepaspectratio]{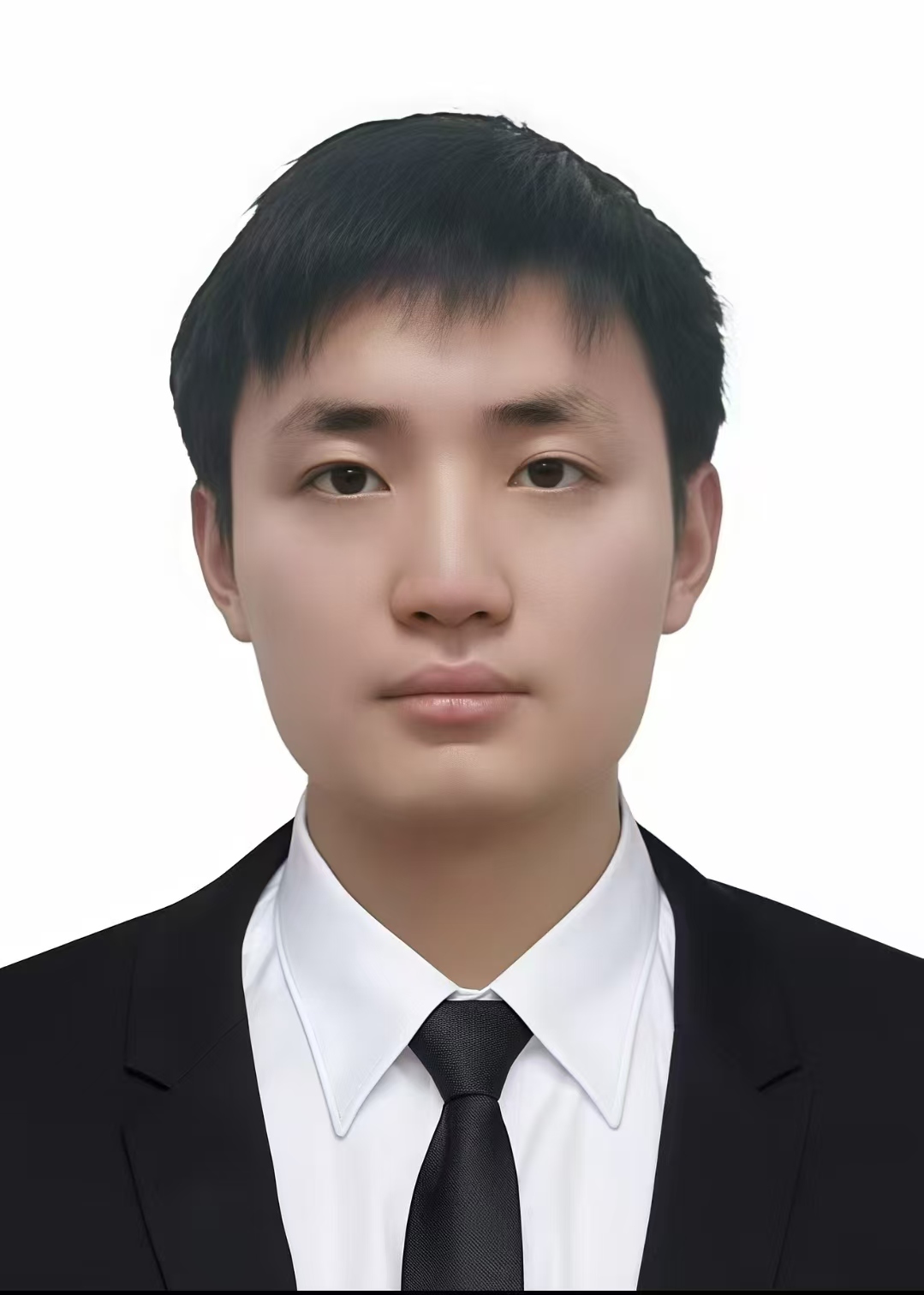}}]{Shuangjun Du}  received the B.E. degree from Sichuan University in 2024. He is currently pursuing the Master's degree with the Institute of Automation, Chinese Academy of Sciences (CASIA). His research interests include computer vision, particularly image and video generation.
\end{IEEEbiography}

\vspace{-12mm}
 \begin{IEEEbiography}[{\includegraphics[width=1in,height=1.25in,clip,keepaspectratio]{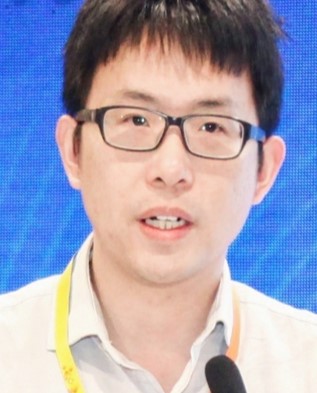}}]{Bo Peng} received the BEng degree from Beihang University and the PhD degree from the Institute of Automation Chinese Academy of Sciences (CASIA), in 2013 and 2018, respectively. Since 2018, he has joined CASIA where he is currently an Associate Professor. His research focuses on computer vision, image forensics, deepfake detection, and responsible AIGC generation. He is the secretary of IEEE Beijing Biometrics Council Chapter and served as a member in several IEEE R10 committees.
\end{IEEEbiography}

\vspace{-12mm}
 \begin{IEEEbiography}[{\includegraphics[width=1in,height=1.25in,clip,keepaspectratio]{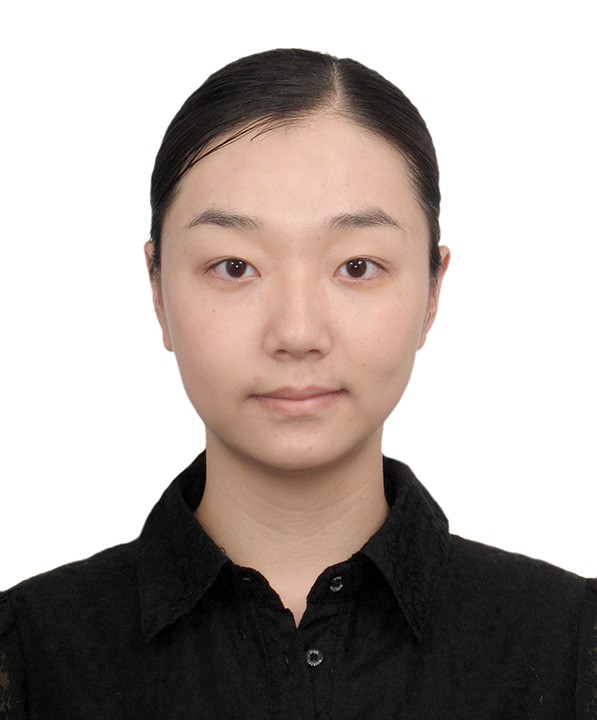}}]{Jing Dong} recieved the PhD degree in Pattern Recognition from the Institute of Automation, Chinese Academy of Sciences, China, in 2010. Since then, she joined the Institute of Automation, Chinese Academy of Sciences and she is currently a Professor. Her research interests include pattern recognition, image processing and digital image forensics including digital watermarking, steganalysis and tampering detection. She also has served as the deputy general of Chinese Association for
 Artificial Intelligence.
 \end{IEEEbiography}

\vspace{-12mm}
\begin{IEEEbiography}[{\includegraphics[width=1in,height=1.25in,clip,keepaspectratio]{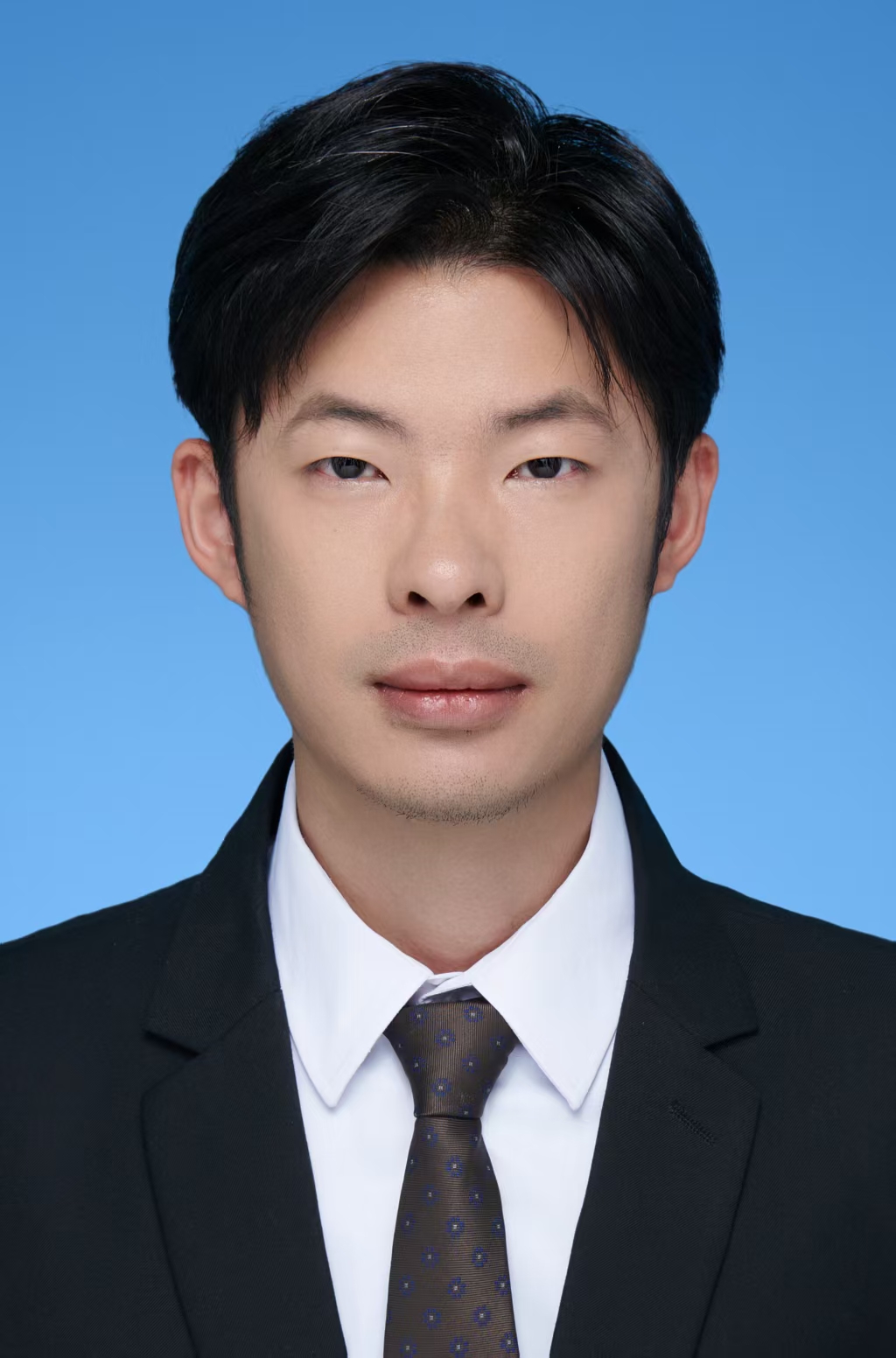}}]{Kun Wang} received his Ph.D. from the University of Science and Technology of China and is currently a Postdoctoral Researcher at Nanyang Technological University. His research focuses on large model safety, trustworthy multi-agent systems, and their applications in spatiotemporal analysis and medicine. He has published over 80 CCF-A papers, including 37 in top-tier venues such as ICLR, NeurIPS, ICML, and TPAMI, with 10+ recognized as Oral or Spotlight presentations. Dr. Wang is a recipient of the ICLR 2025 Best Paper Award. He serves as an Area Chair for ICLR and ICML and regularly reviews for leading AI conferences and journals.
\end{IEEEbiography}

\vspace{-12mm}
\begin{IEEEbiography}[{\includegraphics[width=1in,height=1.25in,clip,keepaspectratio]{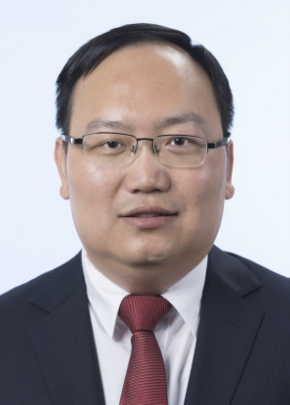}}]{Zhenan Sun}
  received the Ph.D. degree from the Institute of Automation, Chinese Academy of Sciences (CASIA) in 2006.
  He is a professor at New Laboratory of Pattern Recognition (NLPR), State Key Laboratory of Multimodal Artificial Intelligence Systems (MAIS), CASIA.
  His current research interests include biometrics, pattern recognition, and computer vision. He is a fellow of the IAPR, and an Associate Editor of the IEEE Transactions on Biometrics, Behavior, and Identity Science.
\end{IEEEbiography}

\vspace{-12mm}
\begin{IEEEbiography}[{\includegraphics[width=1in,height=1.25in,clip,keepaspectratio]{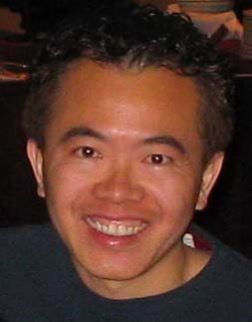}}]{Ming-Hsuan Yang}
is with the Department of Computer Science and Engineering at the University of California, Merced. 
He received Best Paper Award at ICML in 2024, Longuet-Higgins Prize in 2023, Best Paper Honorable Mention at CVPR 2018, the NSF CAREER Award in 2012, and the Google Faculty Award in 2009. 
He is a Fellow of IEEE, ACM, AAAI, and AAAS.
\end{IEEEbiography}

\end{document}